%% file: 4652_camera_ready.tex
\documentclass[runningheads]{llncs}
\usepackage{graphicx}

\usepackage{tikz}
\usepackage{comment}
\usepackage{amsmath,amssymb} 
\usepackage{color}

\usepackage{booktabs}
\usepackage{microtype}
\usepackage{soul}

\usepackage[accsupp]{axessibility}  

\usepackage[pagebackref,breaklinks,colorlinks]{hyperref}

\usepackage{array}
\newcolumntype{H}{>{\setbox0=\hbox\bgroup}c<{\egroup}@{}}

\begin{document}
\pagestyle{headings}
\mainmatter
\def\ECCVSubNumber{4652}  

\title{Neural Scene Decoration\\ from a Single Photograph} 

\titlerunning{Neural Scene Decoration from a Single Photograph}
%
\author{Hong-Wing Pang\inst{1} \and
Yingshu Chen\inst{1} \and
Phuoc-Hieu Le\inst{2} \and
Binh-Son Hua\inst{2} \and \\
Duc Thanh Nguyen \inst{3} \and 
Sai-Kit Yeung\inst{1}
}
\authorrunning{Pang et al.}
%
\institute{Hong Kong University of Science and Technology \and
VinAI Research, Vietnam 
\and
Deakin University
} 
\maketitle

\begin{abstract}
Furnishing and rendering indoor scenes has been a long-standing task for interior design, where artists create a conceptual design for the space, build a 3D model of the space, decorate, and then perform rendering. Although the task is important, it is tedious and requires tremendous effort. In this paper, we introduce a new problem of domain-specific indoor scene image synthesis, namely neural scene decoration. Given a photograph of an empty indoor space and a list of decorations with layout determined by user, we aim to synthesize a new image of the same space with desired furnishing and decorations. Neural scene decoration can be applied to create conceptual interior designs in a simple yet effective manner. Our attempt to this research problem is a novel scene generation architecture that transforms an empty scene and an object layout into a realistic furnished scene photograph. We demonstrate the performance of our proposed method by comparing it with conditional image synthesis baselines built upon prevailing image translation approaches both qualitatively and quantitatively. We conduct extensive experiments to further validate the plausibility and aesthetics of our generated scenes. Our implementation is available at \url{https://github.com/hkust-vgd/neural_scene_decoration}.

\end{abstract}

\input{sections/intro}
\input{sections/related}
\input{sections/method}
\input{sections/experiments}
\input{sections/conclusion}

\noindent\textbf{Acknowledgment.} This paper was partially supported by an internal grant from HKUST (R9429) and the HKUST-WeBank Joint Lab.

\clearpage
%
%
\bibliographystyle{splncs04}
\bibliography{egbib}

\clearpage 

\noindent\textbf{\Large Supplementary Material}
\appendix
\begin{abstract}
	In this supplementary material, we first present more qualitative results with additional analysis on the diversity, background impact, an additional comparison to text-guided image synthesis (Section~\ref{sec:qualitative_results}). We then provide an additional ablation study on our discriminator design (Section~\ref{sec:ablation}) along with complementary details of our user study (Section~\ref{sec:user}). Finally, we describe in detail our proposed architecture (Section~\ref{sec:architecture}) and its implementation (Section~\ref{sec:implementation}). 
\end{abstract}

\section{Qualitative Results}
\label{sec:qualitative_results}

\subsection{Diversity of generated images}
Our method achieves diversity in the following ways. 
First, the input layout controls the output diversity. One can change the input layout to change how the scene is decorated. 
Second, given the same background and layout,  diversity in the appearance of scene objects can still be obtained.
Technically, this is achieved by changing the initial latent code of the generator and finetune the generator. 

A current limitation is that our model ignores the noise vector, limiting diversity. This problem is also reported in pix2pix~\cite{isola2017image}. 
Revising our network architecture for greater diversity would be a future work, e.g., use the noise injection by OASIS~\cite{schonfeld2021oasis}.

\begin{figure*}[h]
	\begin{center}
		
		\resizebox{0.8\linewidth}{!}{
			
			\begin{tabular}{c c c c}
				\includegraphics[width = 1.25in]{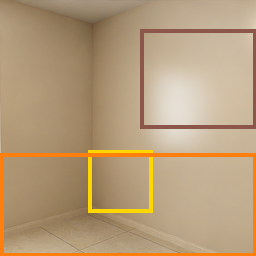} &
				\includegraphics[width = 1.25in]{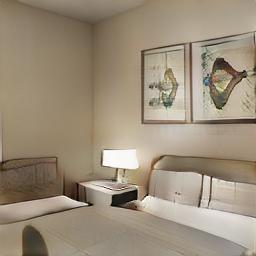} &
				\includegraphics[width = 1.25in]{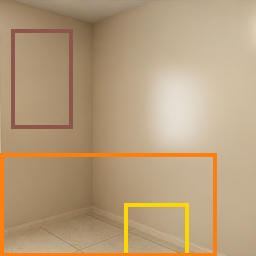} &
				\includegraphics[width = 1.25in]{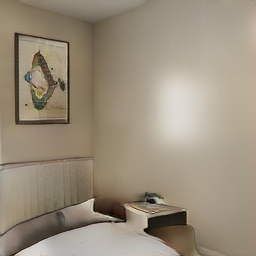} \\
				\includegraphics[width = 1.25in]{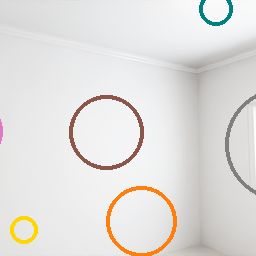} &
				\includegraphics[width = 1.25in]{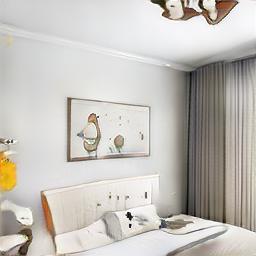} &
				\includegraphics[width = 1.25in]{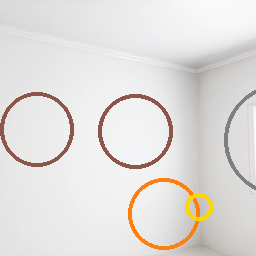} &
				\includegraphics[width = 1.25in]{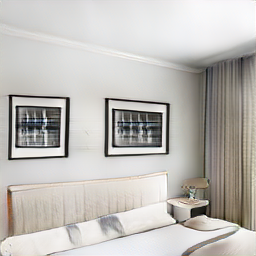} \\
				Input 1 & Output 1 & Input 2 & Output 2 \\
				
				\multicolumn{4}{c}{
					\includegraphics[width = 0.08in]{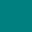} \small{lamp} \quad 
					\includegraphics[width = 0.08in]{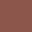} \small{picture} \quad
					\includegraphics[width = 0.08in]{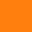} \small{bed} \quad 
					\includegraphics[width = 0.08in]{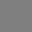} \small{curtain} \quad
					\includegraphics[width = 0.08in]{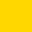} \small{nightstand} \quad
					\includegraphics[width = 0.08in]{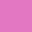} \small{desk} \quad
				}\\
				
			\end{tabular}
			
		} 
		
	\end{center}
	\caption{Diversity evaluation. Generation results under same background image $X$ with different object layouts.}
	\label{fig:diff_layouts}
\end{figure*}

\paragraph{Same background with different layouts.}
We show qualitative results of our method using different layouts on the same background image, demonstrating the diversity of generated contents. Several results of this experiment are presented in Figure~\ref{fig:diff_layouts}.
As can be seen, our model can provide plausible renderings given different layouts. 

\begin{figure}[h]
	\centering
	\includegraphics[width=\textwidth]{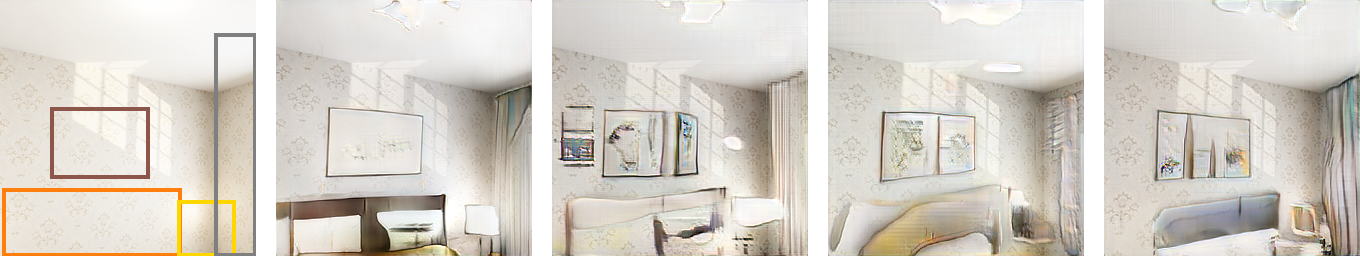}
	\caption{Diversity evaluation. Generation results from same background image $X$ with different model weights.}
	\label{fig:diff_appearance}
\end{figure}

\paragraph{Same background with same layouts.} 
Here we show results of our method using the same background and layout. The diversity is now controlled by the initial latent code of the generator. The results are presented in Figure~\ref{fig:diff_appearance}.
As can be seen, our model can provide plausible diversity in the object appearance.

\subsection{Impact of background to furniture generation.} 
\begin{figure}[h]
	\centering
	\includegraphics[width=\textwidth]{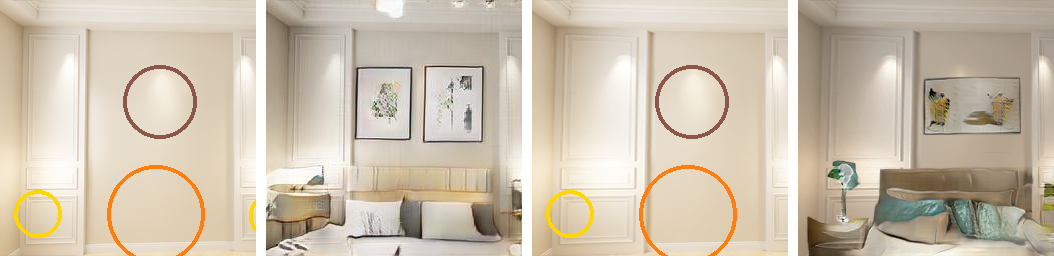}
	\caption{Impact of background to furniture generation.}
	\label{fig:background_impact}
\end{figure}

We analyze the effect of the background to the generation of the furniture. The diagram below shows a simple example where we modify the background image by enlarging the left white backdrop. In Figure~\ref{fig:background_impact}, we see that objects like paintings can conform to this structural change in the background. Other objects like beds only have appearance change. The quantization of the impact of background images is left for future work.

\subsection{Iterative decoration}
\begin{figure}[h]
	\centering
	\includegraphics[width=\textwidth]{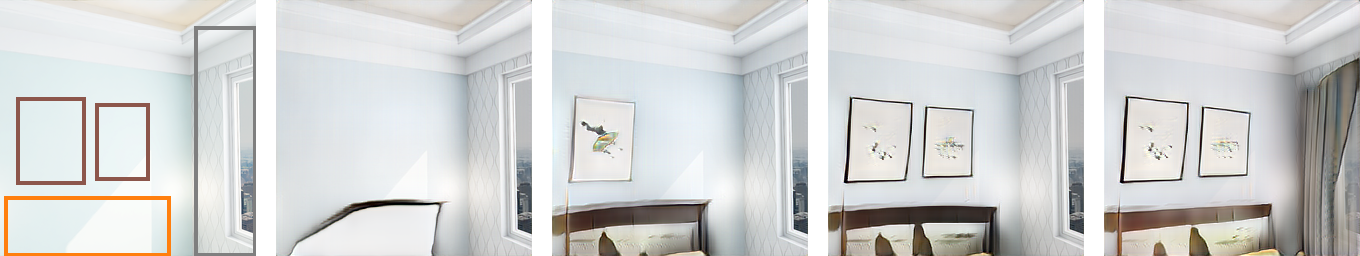}
	\caption{Generation results by adding the objects one at a time.}
	\label{fig:iterative}
\end{figure}

Our model is designed for inferring the decoration at once. While not designed for iterative object insertion, our method can add one object at a time to a limited extent, thanks to the diversity of the scenes and images in the dataset, as shown in the example in Figure~\ref{fig:iterative}. In future work, we could consider using object removal with image inpainting to augment the training data.


\subsection{Comparisons with text-guided image synthesis}
\begin{figure}[h]
	\centering
	\def\sc{0.2}
	\begin{tabular}{c c c c}
		
		\includegraphics[width=\sc\linewidth]{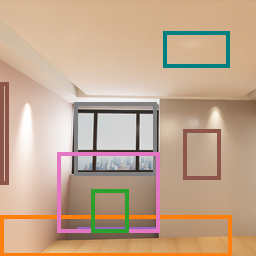}&
		\includegraphics[width=\sc\linewidth]{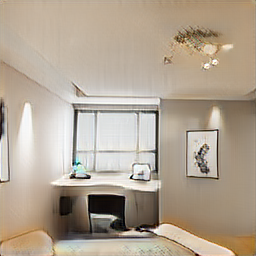}&
		\includegraphics[width=\sc\linewidth]{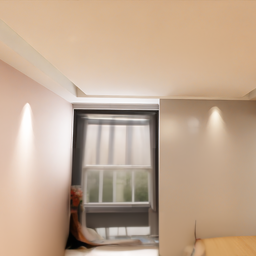}&
		\includegraphics[width=\sc\linewidth]{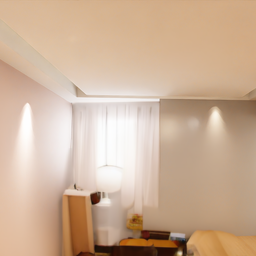} 
		\\
		\includegraphics[width=\sc\linewidth]{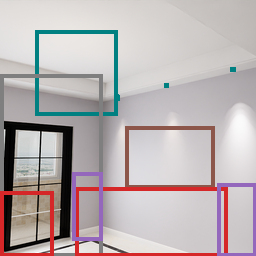}&
		\includegraphics[width=\sc\linewidth]{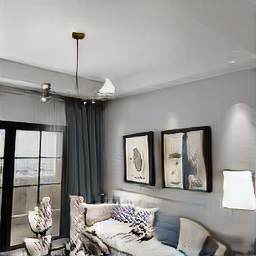}&
		\includegraphics[width=\sc\linewidth]{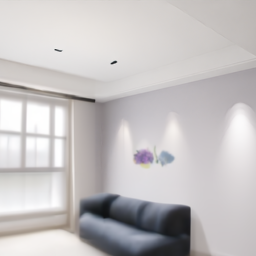}&
		\includegraphics[width=\sc\linewidth]{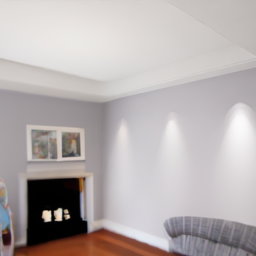}
		\\
		Input & Ours & GLIDE-iter & GLIDE\\
		
	\end{tabular}
	
	\caption{Comparison to text-guided image synthesis method GLIDE~\cite{nichol2021glide}.}
	\label{fig:glide}
\end{figure}

We provide an additional comparison of our method with text-guided image synthesis, which also use coarse layout descriptions similar to ours, unlike fine-grained semantic maps. For text-guided methods, we chose GLIDE~\cite{nichol2021glide} and generated objects by masking target regions and providing a text prompt for each object. 
Specifically, we used released GLIDE (filtered) model for image inpainting in a masked region conditioned on a text prompt. 
We generate objects one by one iteratively via masking each object box with target object text to realize semantic spatially generation (Fig.~\ref{fig:glide} GLIDE-iter column). 
We also inpaint all areas of same boxes of given empty scene once with one text prompt (Fig.~\ref{fig:glide} GLIDE column).
Compared to our method, GLIDE failed to preserve the background (e.g., windows) properly while the generated objects are unaware of the context, making their results not semantically consistent, e.g., the fireplace in the last image.
Additionally, GLIDE takes 4 to 8 seconds to inpaint a 256$\times$256 image which is much slower than our method. 

\subsection{Visual results of more scenes}
\begin{figure*}[h!]
	\def\sc{0.155}
	\begin{center}
		\resizebox{\linewidth}{!}{
			\begin{tabular}{c c c c c}
				\includegraphics[width = 1.23in]{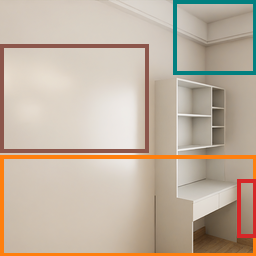} &
				\includegraphics[width = 1.23in]{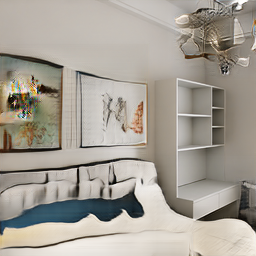} &
				\includegraphics[width = 1.23in]{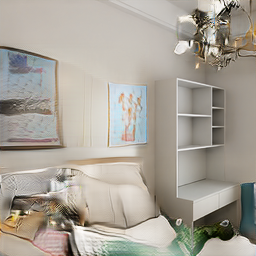} &
				\includegraphics[width = 1.23in]{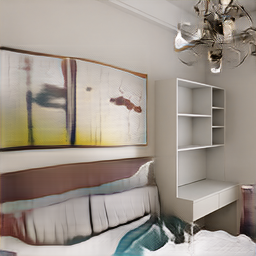} &
				\includegraphics[width = 1.23in]{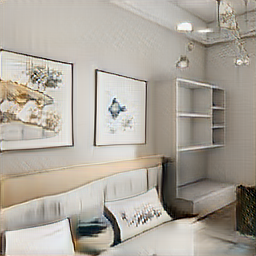} \\
				\includegraphics[width = 1.23in]{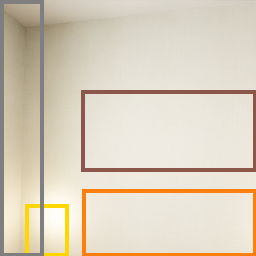} &
				\includegraphics[width = 1.23in]{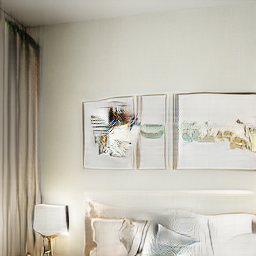} &
				\includegraphics[width = 1.23in]{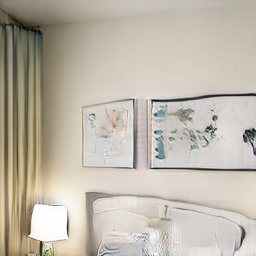} &
				\includegraphics[width = 1.23in]{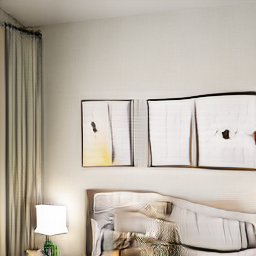} &
				\includegraphics[width = 1.23in]{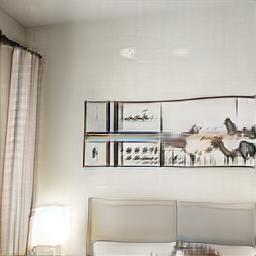} \\
				\includegraphics[width = 1.23in]{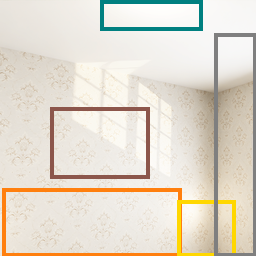} &
				\includegraphics[width = 1.23in]{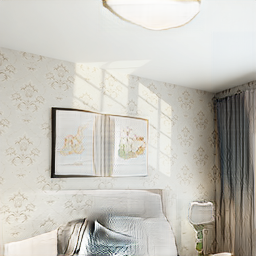} &
				\includegraphics[width = 1.23in]{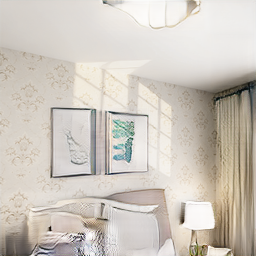} &
				\includegraphics[width = 1.23in]{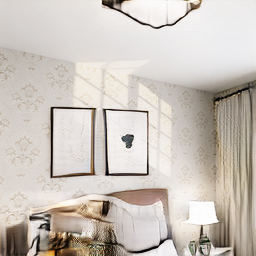} &
				\includegraphics[width = 1.23in]{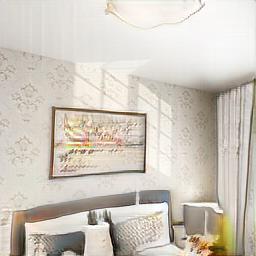} \\
				\includegraphics[width = 1.23in]{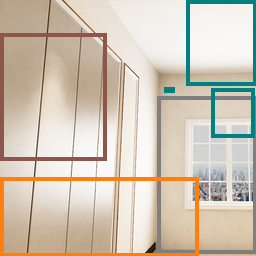} &
				\includegraphics[width = 1.23in]{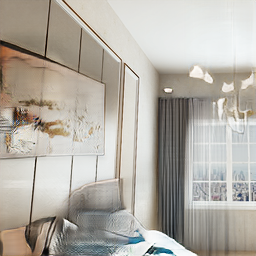} &
				\includegraphics[width = 1.23in]{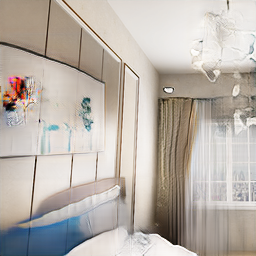} &
				\includegraphics[width = 1.23in]{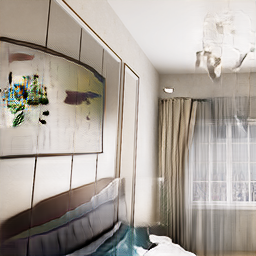} &
				\includegraphics[width = 1.23in]{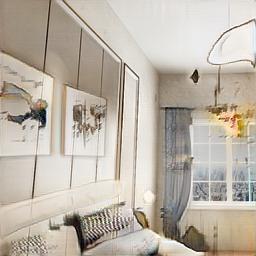} \\
				\includegraphics[width = 1.23in]{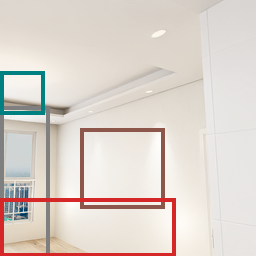} &
				\includegraphics[width = 1.23in]{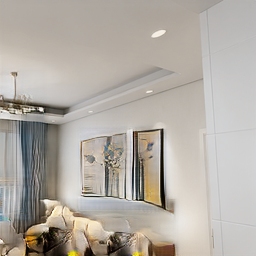} &
				\includegraphics[width = 1.23in]{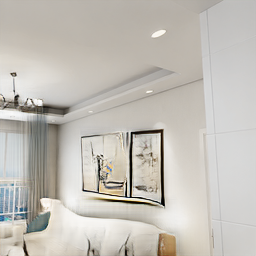} &
				\includegraphics[width = 1.23in]{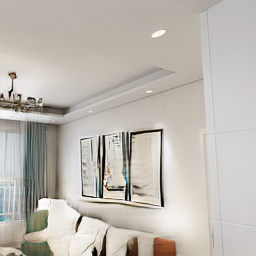} &
				\includegraphics[width = 1.23in]{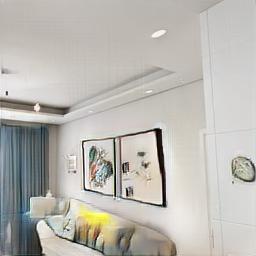} \\
				\includegraphics[width = 1.23in]{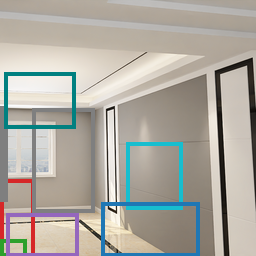} &
				\includegraphics[width = 1.23in]{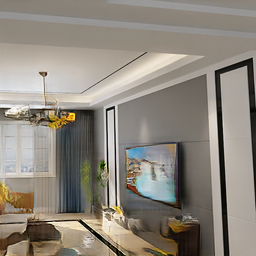} &
				\includegraphics[width = 1.23in]{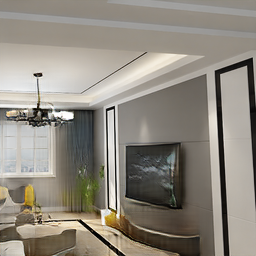} &
				\includegraphics[width = 1.23in]{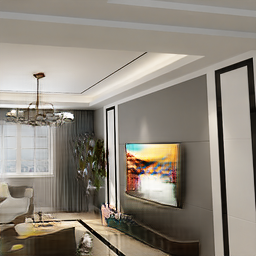} &
				\includegraphics[width = 1.23in]{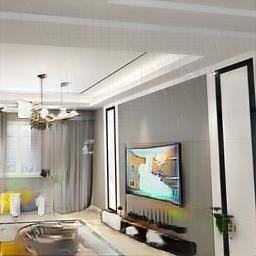} \\
				Input & BachGAN & SPADE & He et. al. & Ours \\
				
				\multicolumn{5}{c}{
					\includegraphics[width = 0.08in]{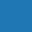} \small{cabinet} \quad 
					\includegraphics[width = 0.08in]{figures/palette/class_lamp.png} \small{lamp} \quad 
					\includegraphics[width = 0.08in]{figures/palette/class_picture.png} \small{picture} \quad
					\includegraphics[width = 0.08in]{figures/palette/class_bed.png} \small{bed} \quad 
					\includegraphics[width = 0.08in]{figures/palette/class_curtain.png} \small{curtain} \quad
				}
				\\
				\multicolumn{5}{c}{
					\includegraphics[width = 0.08in]{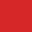} \small{sofa} \quad
					\includegraphics[width = 0.08in]{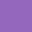} \small{table} \quad
					\includegraphics[width = 0.08in]{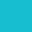} \small{television} \quad 
					\includegraphics[width = 0.08in]{figures/palette/class_nightstand.png} \small{nightstand} \quad
					\includegraphics[width = 0.08in]{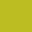} \small{pillow}
				}
			\end{tabular}
		}
	\end{center}
	\caption{Additional generation results for box label format. 
	}
	\label{fig:additional_boxes}
\end{figure*}

\begin{figure*}[h!]
	\def\sc{0.155}
	\begin{center}
		\resizebox{\linewidth}{!}{
			\begin{tabular}{c c c c c}
				\includegraphics[width = 1.23in]{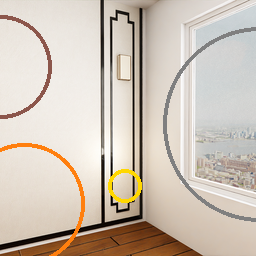} &
				\includegraphics[width = 1.23in]{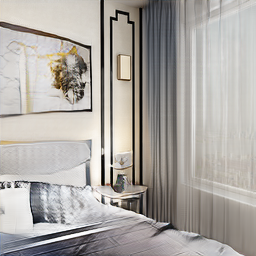} &
				\includegraphics[width = 1.23in]{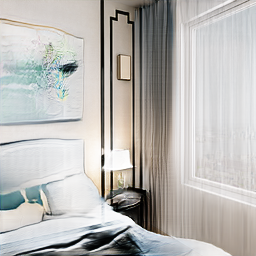} &
				\includegraphics[width = 1.23in]{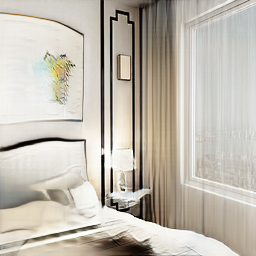} &
				\includegraphics[width = 1.23in]{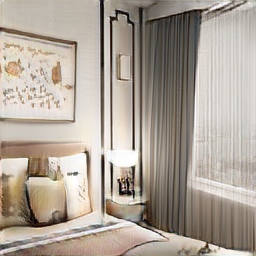} \\
				\includegraphics[width = 1.23in]{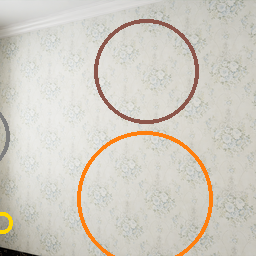} &
				\includegraphics[width = 1.23in]{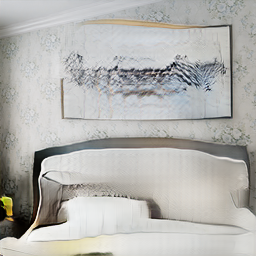} &
				\includegraphics[width = 1.23in]{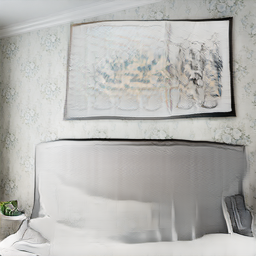} &
				\includegraphics[width = 1.23in]{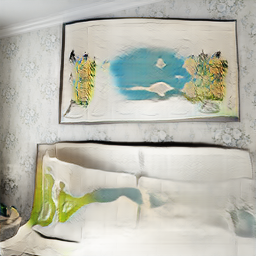} &
				\includegraphics[width = 1.23in]{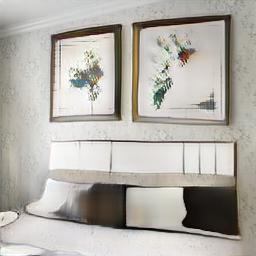} \\
				\includegraphics[width = 1.23in]{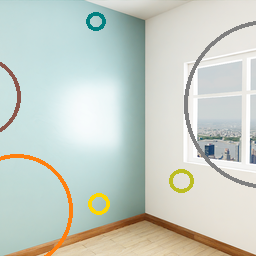} &
				\includegraphics[width = 1.23in]{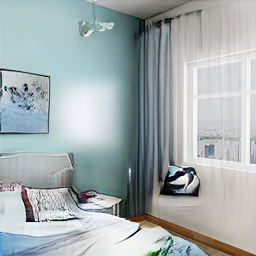} &
				\includegraphics[width = 1.23in]{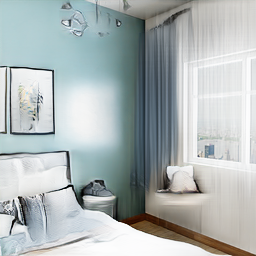} &
				\includegraphics[width = 1.23in]{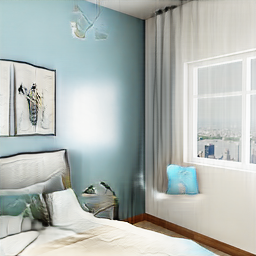} &
				\includegraphics[width = 1.23in]{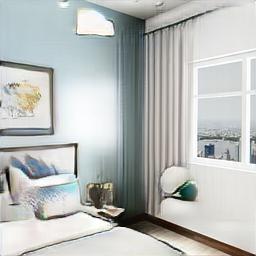} \\
				\includegraphics[width = 1.23in]{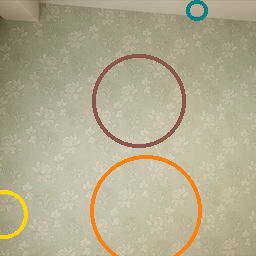} &
				\includegraphics[width = 1.23in]{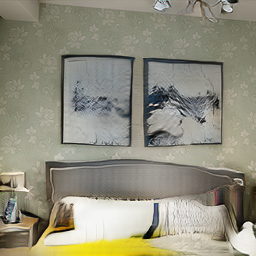} &
				\includegraphics[width = 1.23in]{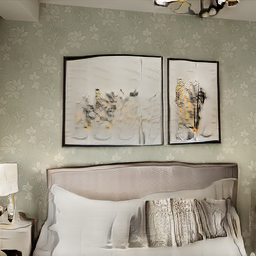} &
				\includegraphics[width = 1.23in]{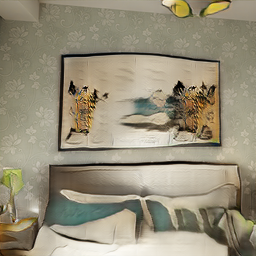} &
				\includegraphics[width = 1.23in]{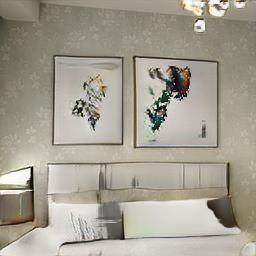} \\
				\includegraphics[width = 1.23in]{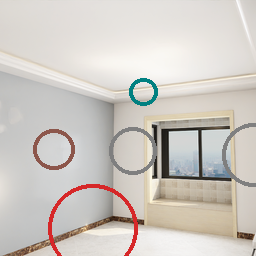} &
				\includegraphics[width = 1.23in]{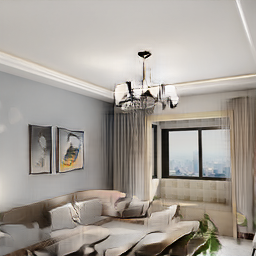} &
				\includegraphics[width = 1.23in]{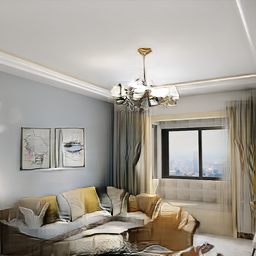} &
				\includegraphics[width = 1.23in]{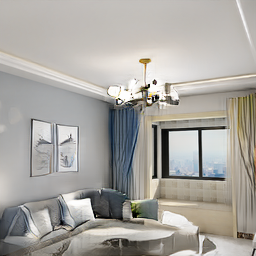} &
				\includegraphics[width = 1.23in]{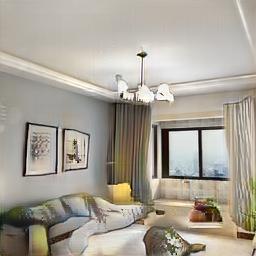} \\
				\includegraphics[width = 1.23in]{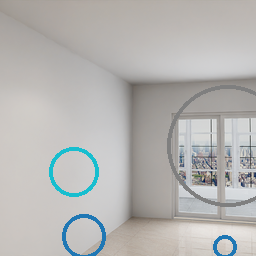} &
				\includegraphics[width = 1.23in]{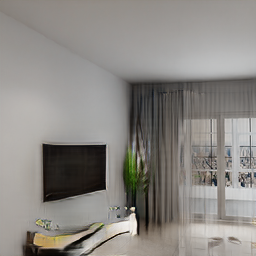} &
				\includegraphics[width = 1.23in]{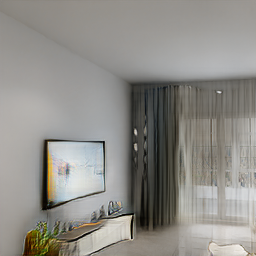} &
				\includegraphics[width = 1.23in]{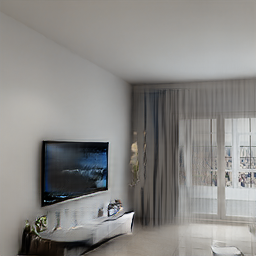} &
				\includegraphics[width = 1.23in]{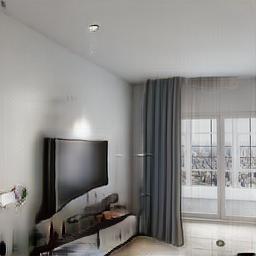} \\
				Input & BachGAN & SPADE & He et. al. & Ours \\
				
				\multicolumn{5}{c}{
					\includegraphics[width = 0.08in]{figures/palette/class_cabinet.png} \small{cabinet} \quad 
					\includegraphics[width = 0.08in]{figures/palette/class_lamp.png} \small{lamp} \quad 
					\includegraphics[width = 0.08in]{figures/palette/class_picture.png} \small{picture} \quad
					\includegraphics[width = 0.08in]{figures/palette/class_bed.png} \small{bed}
					\quad
					\includegraphics[width = 0.08in]{figures/palette/class_curtain.png} \small{curtain} \quad
				}
				\\
				\multicolumn{5}{c}{
					\includegraphics[width = 0.08in]{figures/palette/class_sofa.png} \small{sofa} \quad
					\includegraphics[width = 0.08in]{figures/palette/class_table.png} \small{table} \quad
					\includegraphics[width = 0.08in]{figures/palette/class_television.png} \small{television} \quad 
					\includegraphics[width = 0.08in]{figures/palette/class_nightstand.png} \small{nightstand} \quad
					\includegraphics[width = 0.08in]{figures/palette/class_8.png} \small{pillow}
				}
			\end{tabular}
		} 
	\end{center}
	\caption{Additional generation results for point label format. 
	}
	\label{fig:additional_points}
\end{figure*}

We provide more qualitative results of our method and all baselines (SPADE~\cite{park2019SPADE}, BachGAN~\cite{li2020BachGAN}, He et al.~\cite{he2021context}) in Figure~\ref{fig:additional_boxes} and Figure~\ref{fig:additional_points}. In general, we visually found that bedroom images are often generated in higher quality compared with living room images. This is because bedroom scenes have less variation in their structure, and there are typically less objects decorated in the scenes, leading to lower complexity in scene generation compared to living room scenes. Additionally, we observed that box label format shows more advantages in generating small and relatively fixed size objects. Point label format, on the other hand, allows flexibility in determining the object size and thus works well with large and shape-variable objects.

\section{Ablation Study}
\label{sec:ablation} 
In addition to the ablation study provided in the main paper, here we further explain our discriminator in detail. 
In the main paper, we take the generated image $Y$ as input to the discriminator. 
This is known as the unconditional discriminator as it does not depend on the input $X$.
In fact, image translation methods like pix2pix~\cite{isola2017image}, pix2pixHD~\cite{wang2018high} and SPADE~\cite{park2019SPADE} showed that a conditional version of the discriminator can have better image fidelity. Particularly, the conditional discriminator takes a channel-wise concatenation of the original and generated image (background $X$ and generated image $Y$ in our case) as input to the discriminator. 
Here we provide an experiment to compare the use of unconditional and conditional discriminator in our case. 
Comparison results are reported in Table~\ref{table:concat_ablation}. As shown in the results, the unconditional discriminator has better results in most cases. 
The major difference between our method and image translation methods lies in our data, where the domain gap between the background and the decorated scene is less significant compared to data tested in image translation methods, i.e., sketch or semantic maps vs. real images. 
Therefore, we adopted the unconditional discriminator in our work.

\begin{table*}[h]
	\scriptsize
	\begin{center}
		\begin{tabular}{l | c c | c c | c c | c c} 
			\toprule
			& \multicolumn{4}{c}{Bedroom} & \multicolumn{4}{c}{Living room} \\
			& \multicolumn{2}{c}{Boxes} & \multicolumn{2}{c}{Points} & \multicolumn{2}{c}{Boxes} & \multicolumn{2}{c}{Points}\\
			\midrule
			Discriminator & FID $\downarrow$ & KID{\scriptsize$\times10^3$} $\downarrow$ & FID $\downarrow$ & KID{\scriptsize$\times10^3$} $\downarrow$ & FID $\downarrow$ & KID{\scriptsize$\times10^3$} $\downarrow$ & FID $\downarrow$ & KID{\scriptsize$\times10^3$} $\downarrow$\\
			\midrule
			Without $X$ & 20.596 & \textbf{11.609} & \textbf{15.108} & \textbf{6.797} & \textbf{18.478} & \textbf{10.113} & \textbf{17.986} & \textbf{9.421} \\
			With $X$ & \textbf{20.511} & 12.490 & 16.038 & 12.564 & 21.640 & 14.850 & 21.653 & 14.974 \\
			\bottomrule
		\end{tabular}
	\end{center}
	\caption{Comparison between unconditional discriminator (without $X$) and conditional discriminator (with $X$).}
	\label{table:concat_ablation}
\end{table*}

\section{User Study}
\label{sec:user} 
Our user study has 26 participants; each participant was asked with 48 questions. For each question, we presented two decorated images, one image was generated with our method and the other one was generated by a baseline. Both images were generated from the same input scene. We asked each participant to choose the image that they considered to be more natural and realistic. 
Each question belongs to one of 12 test settings, which is a combination of the following factors: 3 baselines to compare, 2 label formats (box label / point label), and 2 test cases (bedroom / living room). We randomly picked 4 samples for each setting, i.e., each participant was presented with a total of 48 image pairs in random order. The order that two images in a pair were shown in each question was also made randomly.

In general, our model is often preferred on images generated with point label format, especially in the bedroom test case with fewer objects and clutter. When using box label format, our method still produces results with on par quality compared with the baselines.


\clearpage

\clearpage
\section{Network Architecture}
\label{sec:architecture}

\subsection{Generator}

Table \ref{table:generator_arch} describes the input and output dimensions used in the sequence of generator blocks in our generator. For each generator block with $v_i$ input and $v_o$ output channels, the object layout $L$ first modulates the feature map using a SPADE residual block similar \cite{park2019SPADE}, which consists of two consecutive SPADE layers with ReLU activations, as well as a skip connection across the block. Unlike \cite{park2019SPADE}, we do not add a convolutional layer after each SPADE layer in the residual blocks. The number of channels remains to be $v_i$ before and after the SPADE block, and the number of hidden channels in SPADE layers is set to $v_i/2$. Following the SPADE block, we upsample the feature map by a factor of 2, pass through a convolutional layer with $2c_o$ output channels, batch norm layer and finally through a gated linear unit (GLU), following the convolutional block implementation in~\cite{liu2021faster}. All aforementioned convolutional layers have a kernel size of 3 and padding size of 1.

The last two generator blocks use the SLE module in~\cite{liu2021faster} to modulate the feature maps with earlier, smaller-resolution feature maps. We pass the output of the source generator block through an adaptive pooling layer to reduce its spatial size to $4\times 4$, then use a $4\times4$ convolutional layer of kernel size of 4 to collapse the spatial dimensions, reducing the feature map to a 1D vector. This is passed through a LeakyReLU (0.1) activation, $1\times1$ convolutional layer and sigmoid function to obtain a 1D vector of size $v_o$, where $v_o$ is the number of output channels of the destination generator block. This vector is multiplied channel-wise with the feature map inside the destination generator block, right after the upsample operation.

\subsection{Discriminator}

The main discriminator $D_{adv}$ consists of five discriminator blocks, followed by an output convolution module. Each discriminator block consists of two sets of convolutional layers. The first set has a kernel size of 4 and stride of 2, and is responsible for downsampling feature maps by a factor of 2. The second one has a kernel size of 3 and padding size of 1, and transforms the feature maps from $v_i$ to $v_o$ channels, where $v_i$ and $v_o$ are the numbers of channels listed in Table \ref{table:discriminator_arch}. The output convolution module downsamples the feature map to $4\times4$, and is followed by a final $4\times 4$ convolution layer reducing the feature map to one single logit. The object layout discriminator $D_{obj}$ takes the $32\times 32$ feature map output from $D_{adv}$ and repeatedly downsamples the feature map by a factor of 2, using convolutional layers with kernel size of 4 and stride of 2. Like $D_{adv}$, the final feature maps are reduced to a single logit via a final convolutional layer.  Each convolutional layer in $D_{adv}$ and $D_{obj}$ - except the final layer - is followed by batch norm and LeakyReLU (0.1) activations.


\begin{table}[t]
	\begin{center}
		\begin{tabular}{c | c c c } 
			\toprule
			Block $\#$ & Resolution & SLE source block & Features \\
			\midrule
			2 & $4\rightarrow8$ & -- & $12\rightarrow512$ \\
			3 & $8\rightarrow16$ & -- & $512\rightarrow512$ \\
			4 & $16\rightarrow32$ & -- & $512\rightarrow256$ \\
			5 & $32\rightarrow64$ & -- & $256\rightarrow128$ \\
			6 & $64\rightarrow128$ & 2 & $128\rightarrow64$ \\
			7 & $128\rightarrow256$ & 3 & $64\rightarrow32$ \\
			\bottomrule
		\end{tabular}
	\end{center}
	
	\caption{List of generator blocks and their properties.}
	\label{table:generator_arch}
\end{table}

\begin{table}[t]
	\begin{center}
		\begin{tabular}{c | c c c } 
			\toprule
			Block $\#$ & Resolution & Features \\
			\midrule
			7 & $256\rightarrow128$ & $3\rightarrow32$ \\
			6 & $128\rightarrow64$ & $32\rightarrow64$ \\
			5 & $64\rightarrow32$ & $64\rightarrow128$ \\
			4 & $32\rightarrow16$ & $128\rightarrow256$ \\
			3 & $16\rightarrow8$ & $256\rightarrow512$ \\
			Output & $8\rightarrow1$ & $512\rightarrow1$ \\
			\midrule
			4 & $32\rightarrow16$ & $64\rightarrow128$ \\
			3 & $16\rightarrow8$ & $128\rightarrow256$ \\
			2 & $8\rightarrow4$ & $256\rightarrow256$ \\
			1 & $4\rightarrow2$ & $256\rightarrow256$ \\
			Output & $2\rightarrow1$ & $256\rightarrow1$ \\
			\bottomrule
		\end{tabular}
	\end{center}
	
	\caption{List of discriminator blocks in $D_{adv}$ and convolution layers in $D_{obj}$.}
	\label{table:discriminator_arch}
\end{table}

\section{Implementation Details}
\label{sec:implementation}

\subsection{Dataset}



As presented in the main paper, the semantic labels for images in the Structured3D dataset are retrieved from the NYU-Depth V2 dataset \cite{Silberman:ECCV12}. Five classes: \texttt{window}, \texttt{door}, \texttt{wall}, \texttt{ceiling}, and \texttt{floor} are considered as ``background'' and appear in both both empty and decorated scenes. The remaining classes represent ``foreground'' and are used in decorated scenes only. In addition, since the distribution of the foreground classes is highly unbalanced, and some classes do not really exist in the Structured3D dataset, only a subset of these foreground classes were used in our experiments. We show the list of the foreground classes used in our work in Table~\ref{table:palette}.

\begin{table}[t]
	\begin{center}
		\begin{tabular}{c c c c} 
			\toprule
			Name & Color & Name & Color\\ 
			\midrule
			\texttt{cabinet} & \includegraphics[width = 0.15in]{figures/palette/class_cabinet.png} & \texttt{picture} & \includegraphics[width = 0.15in]{figures/palette/class_picture.png}\\
			\texttt{bed} & \includegraphics[width = 0.15in]{figures/palette/class_bed.png} & \texttt{curtain} & \includegraphics[width = 0.15in]{figures/palette/class_curtain.png}\\
			\texttt{chair} & \includegraphics[width = 0.15in]{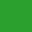} & \texttt{television} & \includegraphics[width = 0.15in]{figures/palette/class_television.png}\\
			\texttt{sofa} & \includegraphics[width = 0.15in]{figures/palette/class_sofa.png} & \texttt{nightstand} & \includegraphics[width = 0.15in]{figures/palette/class_nightstand.png}\\
			\texttt{table} & \includegraphics[width = 0.15in]{figures/palette/class_table.png} & \texttt{lamp} & \includegraphics[width = 0.15in]{figures/palette/class_lamp.png}\\
			\texttt{desk} & \includegraphics[width = 0.15in]{figures/palette/class_desk.png} & \texttt{pillow} & \includegraphics[width = 0.15in]{figures/palette/class_8.png}\\
			\bottomrule
		\end{tabular}
	\end{center}
	\caption{Foreground classes used in our work.}
	\label{table:palette}
\end{table}


We carried out experiments on two subsets of the Structured3D dataset - bedrooms and living rooms, as those sets contain enough samples for training and testing. Note that each scene in the Structured3D dataset is associated with a room type label, that allows us to identify bedroom and living room scenes. To provide enough clue for a scene type, we filtered out images that contain less than 4 objects. For each source image, we resized the image from the original size $1280\times720$ to $456\times256$, then cropped two images with size $256\times256$ from each source image. Images were cropped such that, for foreground object pixels, at least 60\% were still present in cropped regions. We report the total number of training and test samples for each set in Table~\ref{table:datasets}.

\begin{table}[b]
	\begin{center}
		\begin{tabular}{c c c c c} 
			\toprule
			Data split & No training images & No test images \\ 
			\midrule
			Bedroom & 28,038 & 4,931 \\
			Living room & 19,636 & 3,976 \\
			\bottomrule
		\end{tabular}
	\end{center}
	\caption{Statistics of data used for training and testing.}
	\label{table:datasets}
\end{table}


\subsection{Data augmentation}

\begin{figure}[t]
	\centering
	\begin{minipage}{.4\textwidth}
		\centering
		\includegraphics[width = .5\textwidth]{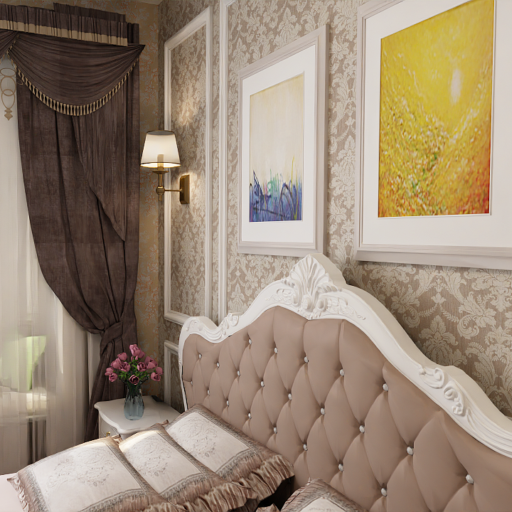}%
		\includegraphics[width = .5\textwidth]{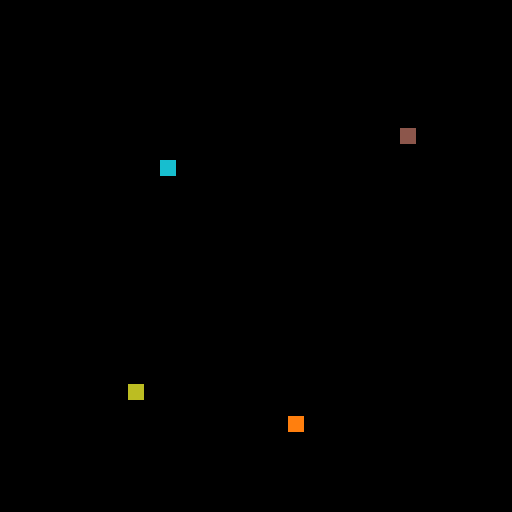}
		\centerline{(a)}
	\end{minipage}
	\begin{minipage}{.4\textwidth}
		\centering
		\includegraphics[width = .5\textwidth]{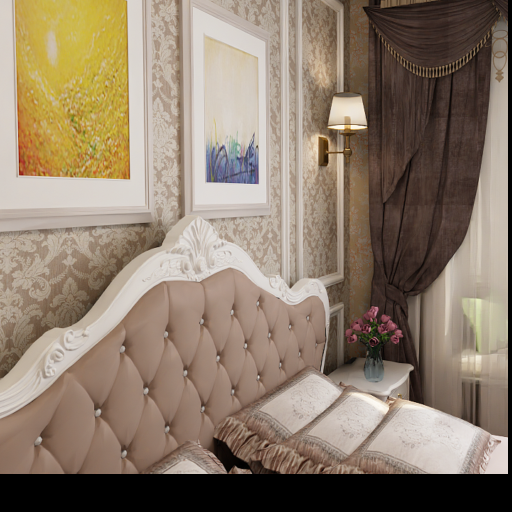}%
		\includegraphics[width = .5\textwidth]{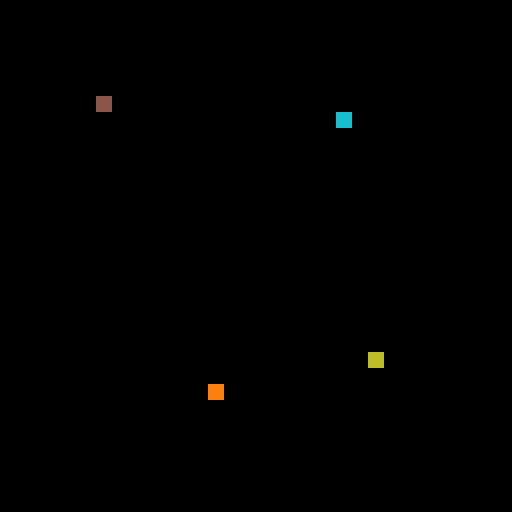}
		\centerline{(b)}
	\end{minipage}
	\caption{(a) Sample image with corresponding object layout map where each dot shows the location and semantic label (via the color) for an object. (b) Same sample after translation and horizontal flipping.}
	\label{fig:augment}
\end{figure}

A direct consequence of training on smaller subsets of the Structured3D dataset is that the number of usable training samples the model observes is greatly reduced. To deal with this issue, we implemented the DiffAugment technique \cite{zhao2020diffaugment} in our training process. DiffAugment improves generation quality by randomly perturbing both the generated and real images with differentiable augmentations when training both $G$ and $D$, and is reported to significantly boost the generation quality of state-of-the-art unconditional StyleGAN2 \cite{karras2020stylegan2,karras2020ada} architecture when training data is limited to a few thousand samples. Thus, we adopt this technique when training on our architecture, in order to compensate for the reduction of training samples.

While the authors of DiffAugment proposed multiple augmentation methods, we only applied translation augmentation to the images. This is because other methods (e.g., random square cutouts) may affect the integrity of decorated scene images. We set the translation augmentation probability to 30\%, and also horizontally flipped the images for 50\% of the time. For each augmented image, its corresponding object layout was also perturbed in the same manner. Figure~\ref{fig:augment} shows an example of our augmentation scheme.



\subsection{Implementation Notes}

While our proposed model can make plausible object locations, we notice that the arrangement of the objects currently lacks flexibility. For example, supplying an object label $L$ with only one to two objects is less likely to result in realistic decorated scenes. This is probably due to the fact that the training dataset only contains fully decorated rooms, and therefore the generator is not trained to produce partially decorated rooms. Likewise, our model also tends to perform fairly on object arrangements that rarely occur in the training dataset.

Additionally, we found that multiple object instances in an image are occasionally labelled by Structured3D with the same object ID, e.g., paintings and curtains. This explains why a single \texttt{picture} object label can result in two (or more) generated paintings. Reflections and highlights caused by foreground objects (e.g., lights) are also present in empty scene images, which could hinder the ability of our approach when generalizing to real-life empty scene images that are not lit up.

\end{document}

%% file: sections/intro.tex
\section{Introduction}

Furnishing and rendering indoor scenes is a common task for interior design. This is typically performed by professional designers who carefully craft a conceptual design and furniture placement, followed by extensive modeling via CAD/CAM software to finally create a realistic image using a powerful rendering engine. Such a task often requires extensive background knowledge and experience in the field of interior design, as well as high-end professional software. This makes it difficult for lay users to design their own scenes from scratch.

On the other hand, different image synthesis methods have been developed and become popular in the field. Various types of deep neural networks - typically in the form of auto-encoders~\cite{kingma2013auto} and generative adversarial networks~\cite{goodfellow2014generative} - have shown their capability of synthesizing images from a variety of inputs, e.g., semantic maps, text description. 

Deep neural networks have also been explored in the generation of indoor scenes. Specifically, an indoor scene can be generated from a collection of objects placed in specific arrangements in 3D space or 2D forms such as top-down coordinates of objects or a floor plane.

In this paper, we combine both research directions: image synthesis and scene generation, into a new task, namely \emph{neural scene decoration}. To solve this task, we propose a scene generation architecture that accepts an empty scene and an object layout as input and produces images of the scene decorated with furniture provided in the object layout (see Figure~\ref{fig:baseline_boxes}). Although several existing image synthesis techniques could be applied to create scenes from simple input, e.g., object bounding boxes, the problem of scene generation from an empty background has not yet been adequately explored. Our goal is to generate images of decorated scenes with improved visual quality, e.g., the placement of objects should be coherent.


An immediate application of neural scene decoration is creating conceptual designs of interior space. In particular, based on a few example images of a scene, a conceptual design of the scene can be made prior to texturing and rendering objects in the scene. Despite the availability of computer-aided design software, creating interior designs is still a challenging task as it requires close collaborations between artists and customers, and may involve third parties. The entire process of making interior designs is also expensive as it requires manual operations, and thus could take several days to complete one design. In this paper, we aim to make neural scene decoration simple and effective in creating realistic furnished scenes. To this end, we have made the following contributions in our work.
\begin{itemize}
\item A new task on scene synthesis and modeling that we name as \emph{neural scene decoration}: synthesize a realistic image with furnished decorations from an empty background image of a scene and an object layout.
\item A neural network architecture that enables neural scene decoration in a simple and effective manner. 
\item Extensive experiments that demonstrate the performance of our proposed method and its potential for future research. Quantitative evaluation results show that our method outperforms previous image translation works. Qualitative results also confirm the ability of our method in generating realistic-looking scenes.
\end{itemize}

%% file: sections/related.tex
\section{Related Work}

\subsection{Image Synthesis}

Prior to the resurgence of deep learning, editing a single photograph has often been done by building a physical model of the scene in the photo for object composition and rendering~\cite{karsch2011legacy,FisherRSFH12,karsch2014composite,karsch2015thesis}. Such an approach requires tremendous effort as it involves huge manual manipulations. 

Deep neural networks, well-known for their learning and generalization capabilities, have been used recently for image synthesis. Technically, network architectures used in existing image synthesis methods fall within two categories: auto-encoders (AEs)~\cite{kingma2013auto} and generative adversarial networks (GANs)~\cite{goodfellow2014generative}. 
Conditional GAN (cGAN)~\cite{mirza2014conditional} is a variant of GANs where generated results are conditioned on input data. A seminal work of cGAN for image synthesis is the image translation method (a.k.a {pix2pix}) in~\cite{isola2017image}, where a cGAN was used to translate an image from one domain to another domain. Much effort has been made to extend the approach in various directions such as generation of high-resolution images~\cite{wang2018high}, greater appearance diversity~\cite{zhu2017toward}, and from arbitrary viewpoints~\cite{tang2019multi}. {CycleGAN} proposed by Zhu et al.~\cite{zhu2017unpaired} extended the image translation to unpaired data (i.e., there are no pairs of images in different domains in training data) by adopting a cycle-consistency loss. Bau et al.~\cite{BauSPWZZ019} allowed users to edit the latent layer in the generator to control the generated content. Recently, Park et al.~\cite{park2019SPADE} proposed spatially-adaptive normalization (SPADE) to modulate intermediate activations, opposed to feeding input data directly into the generator, to strengthen semantic information during the generation process.

Recent developments in GANs led to the generation of high-resolution images~\cite{karras2018progressive} with different styles~\cite{karras2019stylegan,karras2020stylegan2,karras2020ada} and fewer aliasing~\cite{karras2021alias} in the family of StyleGANs~\cite{StyleGAN} especially for human faces. 
For natural and indoor scenes, GANs have also been applied to 2D layout generation. For example, in LayoutGAN~\cite{li2019layoutgan}, a layout was treated as an image and generated in a GAN-style manner. In HiGAN~\cite{yang2019semantic}, latent layers capturing semantic information such as layout, category and lighting were used for scene synthesis. In HouseGAN~\cite{nauata2020house}, planar apartment room layout and furniture layout were generated. 
A class of conditional image synthesis methods focuses on the problem of translating layout to images~\cite{li2020BachGAN,sun2019reconfig,he2021context,li2021locality}.
In these methods, Li et al.~\cite{li2020BachGAN} combined salient object layout and background hallucination GAN (BachGAN)~\cite{li2020BachGAN} for image generation. LostGANs~\cite{sun2019reconfig} explored the reconfigurability of the layouts, but their results mainly have objects on a simple background, which is structurally different from indoor scenes where objects and background must be geometrically aligned. Recent works focus on improving the image generation quality by further using context~\cite{he2021context} and locality~\cite{li2021locality}.
Compared to these works, our method is a conditional image synthesis focused on indoor scenes while the previous works are often tested with images in the wild like those in MS-COCO~\cite{lin2014microsoft} and Visual Genome dataset~\cite{krishna2016visualgenome}. 
The most related work to ours is perhaps BachGAN, where the generation of indoor scene images from layouts are demonstrated but with random backgrounds. 

\subsection{Scene Modeling}

The problem of neural scene decoration is also relevant to the topic of image-based and 3D scene modeling. Earlier works include the spatial arrangement of objects into a 3D scene with spatial constraints, which is often modeled into objective functions that can be solved using optimization techniques~\cite{Germer2009,yu2011make}.
Additional constraints can also be used to model object relations in a scene. 
For example, Henderson et al.~\cite{henderson2017automatic} modeled the relationships among object placement, room type, room shape, and other high-end factors using graphical models. 
Recently, Li et al.~\cite{LiPXCKSTCCZ19} defined object relations in a scene in a hierarchical structure. Ritchie et al.~\cite{ritchie2019fast} proposed to iteratively insert objects into a scene from a top-down view by four different convolutional neural networks (CNNs) capturing the category, location, orientation, and dimension of objects. 
Wang et al.~\cite{wang2019planit} proposed PlanIT, a framework that defines a high-level hierarchical structure of objects before learning to place objects into a scene.
Zhang et al.~\cite{zhang2020deep} proposed a GAN-like architecture to model the distribution of position and orientation of indoor furniture, and to jointly optimize discriminators in both 3D and 2D (i.e., rendered scene images). Several works utilized spatial constraints as priors, e.g., relation graph prior~\cite{wang2019planit,hu2020graph2plan,nauata2020house}, convolution prior~\cite{wang2018deep}, and performed well on spatial organization. 

Our solution for neural scene decoration shares some ideas with scene unfurnishing~\cite{zhang2016emptying} and scene furnishing~\cite{yu2015clutter,zhang2021mageadd,liang2021decorin}.
Instead of utilizing RGBD as input~\cite{zhang2016emptying}, our method takes only a single input photograph of an empty scene and an object layout.
Both ClutterPalette~\cite{yu2015clutter} and MageAdd~\cite{zhang2021mageadd} make 3D scenes by letting users select objects from a synthetic object database. In contrast, our method generates 2D scenes by automatically learning object appearance from training data. 
DecorIn~\cite{liang2021decorin} only predicts decoration locations on walls, while our method really decorates a scene image from a given background and object layout.

%% file: sections/method.tex
\section{Proposed Method}


\subsection{Problem Formulation}
\label{section:problem}

Our goal is to develop a neural scene decoration (NSD) system that produces a decorated scene image $\hat{Y}\in\mathbb{R}^{3\times W\times H}$, given a background image $X\in\mathbb{R}^{3\times W\times H}$, and an object layout $L$ for a list of objects to be added in the scene (see Figure~\ref{fig:baseline_boxes}). Note that both the generated image $\hat{Y}$ and the background image $X$ are captured from the same scene. Ideally, the NSD system should be able to make $\hat{Y}$ realistically decorated with objects specified in $L$, and also assimilate $\hat{Y}$ to the provided background image $X$.

The format of the object layout $L$ is crucial in determining how easy and effective the NSD system is. In SPADE~\cite{park2019SPADE}, synthesized objects are labeled using a pixel-wise fashion. This manner, however, requires detailed labeling which is not effective in describing complicated objects and also takes considerable effort. In our work, we propose to represent $L$ using simple yet effective formats: \textit{box label} and \textit{point label}. Specifically, let $O=\{o_1,...,o_N\}$ be a set of objects added to $X$; these objects belong to $K$ different classes, e.g., chair, desk, lamp, etc. Each object $o_i$ is associated with a class vector $I_i\in \{0,1\}^{K\times1}$ and a layout map $f_i \in \mathbb{R}^{1\times W \times H}$. The class vector $I_i$ is designed such as $I_i(k)=1$ if $k$ is the class ID of $o_i$, and $I_i(k)=0$, otherwise. We define the object layout $L$ as a 3D tensor: $L\in\mathbb{R}^{K\times W \times H}$ as follow:
\begin{align}
\label{eq:object_layout}
L=\sum_{i=1}^N I_i f_i
\end{align}


\paragraph{Box label.} Like BachGAN~\cite{li2020BachGAN}, a box label indicates the presence of an object by its  bounding box. For each object $o_i$, the layout map $f_i$ is constructed by simply filling the entire area of the bounding box of $o_i$ with 1s, and elsewhere with 0s. Mathematically, we define:
\begin{align}
\label{eq:box_label}
f_i(1,x,y)=\begin{cases}  1, & \text{if } (x,y) \in \mathrm{bounding\_box}(o_i) \\                         0, & \text{otherwise}
         \end{cases}
\end{align}

Box label format has the advantage of indicating the boundary where objects should be inserted and allowing finer control over the rough shape of objects.


\paragraph{Point label.} We devise another representation for $f_i$, where each $o_i$ is specified by its center $\mathbf{c}_i=(c_{i,x}, c_{i,y})$ and its size $s_i$ (i.e., a rough estimate of the area of $o_i$). $f_i$ is defined as a heat-map of $\mathbf{c}_i$ and $s_i$ as follow:
\begin{align}
\label{eq:point_label}
f_i(1,x,y)=\exp{\bigg( - \frac{\|(x,y)-(c_{i,x},c_{i,y})\|^2_2}{s_i} \bigg)}
\end{align}


We find this format interesting to explore, as it allows the user to place decorated objects by simply specifying a rough location and size. The exact forms of decorated objects are automatically inferred based on observed training data. 


\subsection{Architecture Design}
\label{section:design}

\begin{figure}[t]
	\centering
	\includegraphics[width=0.48\linewidth]{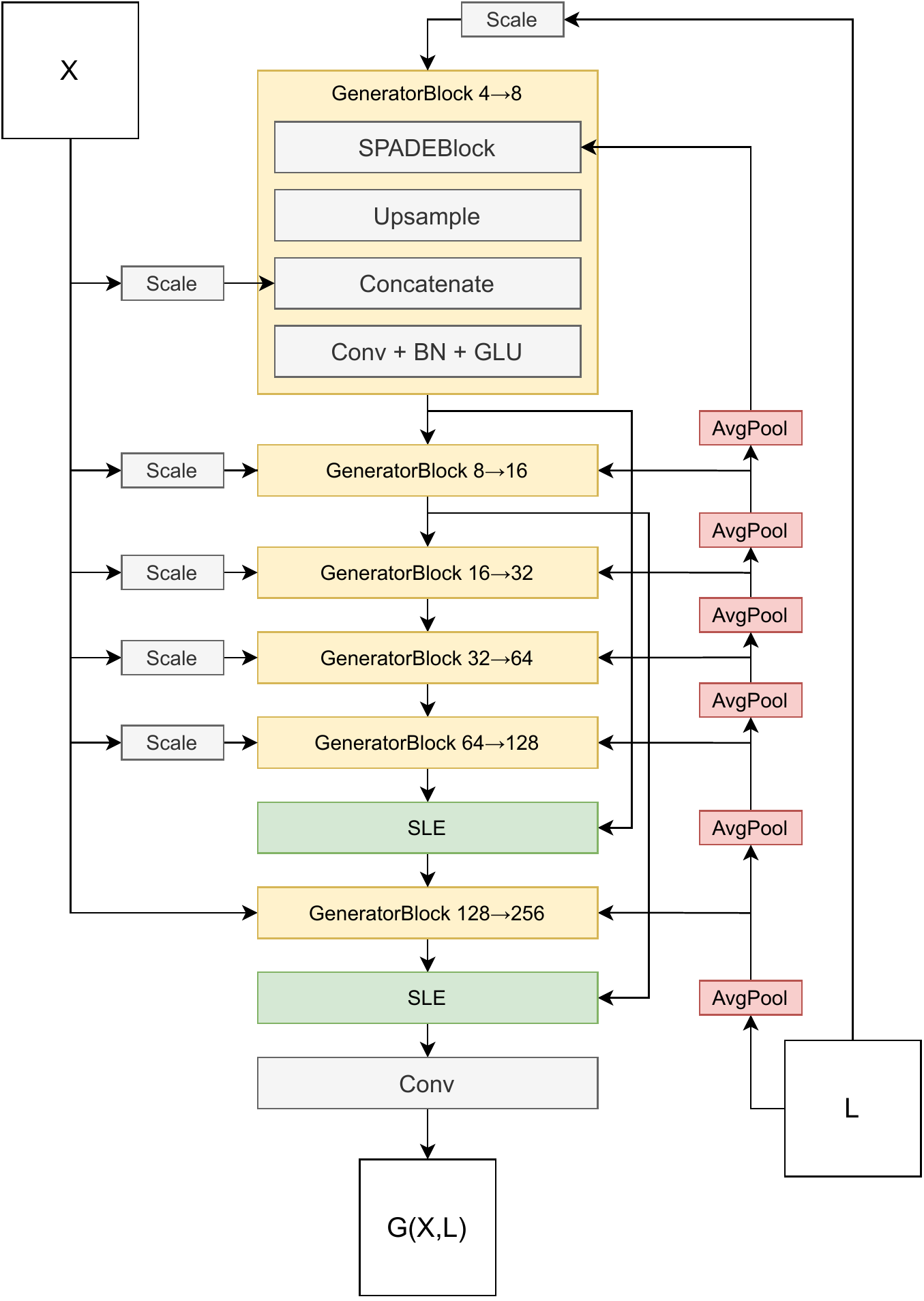} \quad
	\includegraphics[width=0.48\linewidth]{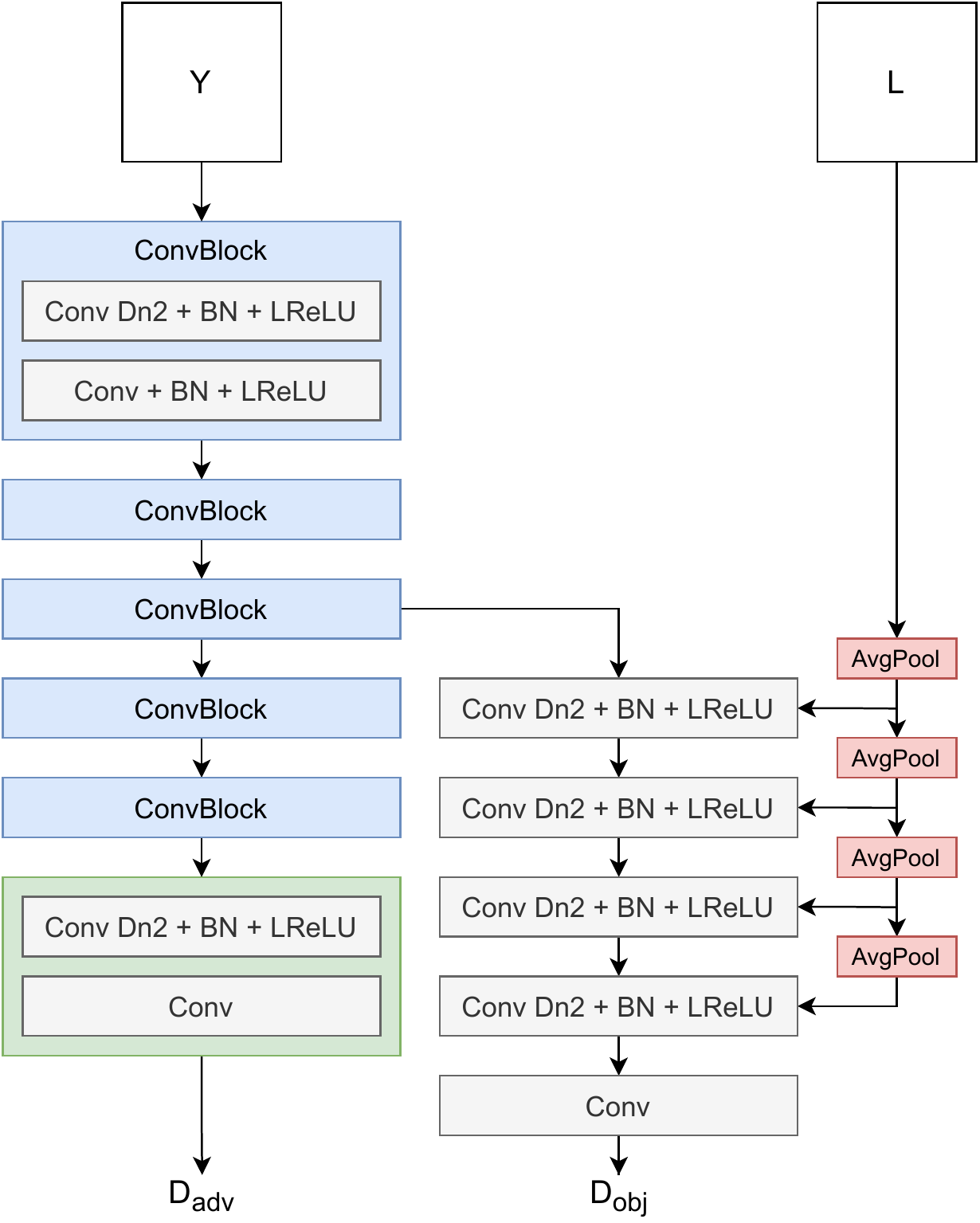}
	\caption{Overview of our generator (left) and discriminator (right). Convolution layers labeled with Dn2 halve the spatial dimensions of input feature maps using stride 2.}
	\label{fig:design}
\end{figure}


Our generator $G(X,L)$ is designed to take a pair of a background image $X$ and an object layout $L$, and generate a decorated scene image $\hat{Y}=G(X,L)$. We also train a discriminator $D(\hat{Y}, L)$ to classify the image $\hat{Y}$ (as synthesized image vs. real image) and to validate if $\hat{Y}$ conforms to the object layout $L$. 



\subsubsection{Generator}

The generator $G(X,L)$ is depicted in Figure~\ref{fig:design}, and includes multiple generator blocks varying from low to high resolutions. For example, the ``GeneratorBlock$4\rightarrow 8$'' generates a feature map of spatial dimension $8\times8$ from $4\times4$.

Each generator block takes input from a feature map produced by its preceding generator (except the first one), an object layout and a background image at a corresponding resolution, and results in a feature map at a higher resolution. Each generator employs a SPADE block~\cite{park2019SPADE} to learn an object layout feature map (used to condition generated content), then up-samples the feature map (by a factor of 2), concatenates it with a down-scaled version of the background image, and finally performs a convolution (Conv) with batch normalization (BN) and GLU. We follow~\cite{liu2021faster} using skip-layer excitations (SLE) that modulates the output of the last two generator blocks with that of the first two blocks. The resulting feature map is passed through a final convolution layer to produce the synthesized image $\hat{Y}$.

To enforce the integration of object layout in the generation process, we insert $L$ into every generator block in a bottom-up manner. Specifically, down-scaled versions of $L$ are created by consecutive average pooling layers and then fused with corresponding feature maps at different resolutions. Likewise, to constrain the consistency in the background of the synthesized image $\hat{Y}$ and the input image $X$, we make $X$ in different scales and insert these scaled images into every generator block. We refer readers to the supplementary material for further details of our architecture.

\subsubsection{Discriminator} 
The discriminator $D(Y, L)$ consists of an adversarial discriminator $D_{adv}(Y)$ and an object layout discriminator $D_{obj}(Y, L)$. We show our discriminator in Figure~\ref{fig:design}.

The adversarial discriminator $D_{adv}$ takes in only the generated image $\hat{Y}=G(X,L)$ as input, and is solely responsible for fitting it towards the target image distribution. In our implementation, $D_{adv}$ is a feedforward network and associated with a global adversarial loss.

We introduce an additional discriminator, $D_{obj}$, to encourage the generated image to follow the object layout provided in $L$. $D_{obj}$ branches off from $D_{adv}$ at the feature map of size 32 $\times$ 32 (see Figure~\ref{fig:design}). In this branch, each feature map is concatenated with an object layout $L_i$ in a proper size $s_i\in\{4,8,16,32\}$. We empirically chose to branch off from $D_{adv}$ at this level as experimental results show that this is sufficient for generating reasonable images. We also found the use of $D_{obj}$ in addition to $D_{adv}$ produces better generation results (see our ablation study).




\subsection{Training}

We train our NSD system using the conventional GAN training procedure, i.e., jointly optimizing both the generator $G$ and discriminator $D$. Specifically, we make use of the hinge adversarial loss function to train $D$ as:
\begin{align}
\mathcal{L}_{D}&=\mathbb{E}_{Y}[\max(0,1+D_{adv}(Y))] \nonumber \\
&+\mathbb{E}_{\hat{Y},L}[\max(0,1-D_{adv}(\hat{Y})-\lambda_{obj}D_{obj}(\hat{Y},L)]
\end{align}
where $Y$ is the decorated scene image paired with $X$ (from training data) and $\hat{Y}=G(X,L)$. $Y$ is also the source image where objects in $L$ are defined. 


The generator $G$ is updated to push the discriminator's output towards the real image direction: 
\begin{align}
\mathcal{L}_G=\mathbb{E}_{X,L}[D(G(X,L))]
\end{align}

In our implementation, we optimize both $\min_{D}\mathcal{L}_D$ and $\min_{G}\mathcal{L}_G$ simultaneously and iteratively. We observed that setting $\lambda_{obj}$ to small values is sufficient for generating plausible results. We set $\lambda_{obj}=0.01$, which empirically works well in our experiments. We trained the discriminator and generator with Adam optimizer. We set the learning rate and batch size to 0.0001 and 32, respectively, and updated the gradients every four iterations. An exponential moving average of the generator weights was applied. The training was conducted on a single RTX 3090 GPU and within 400,000 iterations, which took approximately 44 hours. For inference, the generator performed at 0.760 seconds per image on average. 


%% file: sections/experiments.tex
\section{Experiments}

\subsection{Dataset}

We chose to conduct experiments on the Structured3D dataset~\cite{zheng2020structured3d}, as it is, to the best of our knowledge, the only publicly available dataset with pre-rendered image pairs of empty and decorated scenes. The Structured3D dataset consists of $78,463$ pairs of decorated and empty indoor images, rendered from a total of $3,500$ distinct 3D scenes. Following the recommendation of the dataset authors, we use 3,000 scenes for training and the remaining scenes for validation. 


The Structured3D dataset consists of a variety of indoor scenes, including bedrooms, living rooms, and non-residential locations. However, we carried out experiments on bedrooms and living rooms scenes as those scenes contain sufficient samples for training and testing. 

\subsection{Baselines}


Since we propose a new problem formulation, it is hard to find existing baselines that exactly address the same problem setting. However, our research problem and conditional image synthesis somewhat share a common objective, which is to generate an image conditioned on some given input. Therefore, some image synthesis methods could be adapted and modified to build baselines for comparison. Particularly, we adopted three state-of-the-art image synthesis methods, namely SPADE~\cite{park2019SPADE}, BachGAN~\cite{li2020BachGAN}, and context-aware layout to image generation~\cite{he2021context} for our baselines. Those methods were selected for several reasons. First, they follow conditional image generation setting, i.e., generated contents are conformed to input conditions, e.g., semantic maps in SPADE~\cite{park2019SPADE} and BachGAN~\cite{li2020BachGAN}, and layout structures in~\cite{he2021context,li2021locality,sun2021lostgans}. Second, their input conditions can be customised with minimal modifications to be comparable with ours. We present these modifications below. 

SPADE~\cite{park2019SPADE} is a generative architecture conditioned on pixel-wise semantic maps. 
We concatenate the background image and object layout into each SPADE layer, in place of pixel-wise semantic maps. We also concatenate this tensor to the generated image before passing it to the discriminator. 

BachGAN~\cite{li2020BachGAN} is built upon the SPADE's generator and synthesizes a scene image from foreground object bounding boxes. It uses a \textit{background hallucination module} to make generated background match with the object layout. 
To conform BachGAN to our problem setting, we directly feed the input background image $X$ into the background hallucination module, instead of pooling features from multiple background image candidates. 

We also compare to the context-aware layout to image generation method by He et al.~\cite{he2021context}, which is state-of-the-art in layout-based image generation for general scenes. Their method is reportedly better than LostGANs v2~\cite{sun2021lostgans} in terms of quantitative evaluation, and the implementation is publicly available. We modified their implementation by inserting the background into every block in their generator.

In the supplementary material, we provide an additional comparisons to GLIDE~\cite{nichol2021glide}, a text-guided image synthesis method, which also use coarse layout descriptions similar to ours, unlike fine-grained semantic maps. Please refer to the supplementary material for this result. 

\begin{table*}[t]
\scriptsize
\begin{center}
\begin{tabular}{l | c c | c c | c c | c c} 
\toprule
& \multicolumn{4}{c}{Bedroom} & \multicolumn{4}{c}{Living room} \\
& \multicolumn{2}{c}{Box label} & \multicolumn{2}{c}{Point label} & \multicolumn{2}{c}{Box label} & \multicolumn{2}{c}{Point label}\\
\midrule
Method & FID $\downarrow$ & KID{\scriptsize$\times10^3$}  $\downarrow$ & FID $\downarrow$ & KID{\scriptsize$\times10^3$} $\downarrow$ & FID $\downarrow$ & KID{\scriptsize$\times10^3$} $\downarrow$ & FID $\downarrow$ & KID{\scriptsize$\times10^3$} $\downarrow$\\
\midrule
SPADE~\cite{park2019SPADE} & 23.780 & 12.622 & 20.345 & 9.850 & 21.527 & 12.594 & 19.471 & 10.412 \\
BachGAN~\cite{li2020BachGAN} & 21.319 & 10.054 & 18.829 & 7.932 & 20.463 & 11.961 & 18.997 & 9.446 \\
He et al.~\cite{he2021context} & 21.311 & \textbf{9.869} & 18.899 & 7.958 & 19.732 & 10.828 & 18.762 & 9.656 \\
Ours & \textbf{20.596} & 11.609 & \textbf{15.108} & \textbf{6.797} & \textbf{18.478} & \textbf{10.113} & \textbf{17.986} & \textbf{9.421} \\
\bottomrule
\end{tabular}
\end{center}

\caption{Quantitative assessment of our method against various baselines. Lower FID/KID values indicate better performance.}
\label{table:metrics_points}
\end{table*}

\begin{table*}[t]
\scriptsize
\begin{center}
\begin{tabular}{l | c c | c c | c c | c c} 
\toprule
& \multicolumn{4}{c}{Bedroom} & \multicolumn{4}{c}{Living room} \\
& \multicolumn{2}{c}{Boxes} & \multicolumn{2}{c}{Points} & \multicolumn{2}{c}{Boxes} & \multicolumn{2}{c}{Points}\\
\midrule
Descriptor & FID $\downarrow$ & KID{\scriptsize$\times10^3$} $\downarrow$ & FID $\downarrow$ & KID{\scriptsize$\times10^3$} $\downarrow$ & FID $\downarrow$ & KID{\scriptsize$\times10^3$} $\downarrow$ & FID $\downarrow$ & KID{\scriptsize$\times10^3$} $\downarrow$\\
\midrule
$D_{adv}$ & 22.341 & 13.245 & 21.069 & 11.231 & 25.051 & 15.950 & 24.786 & 14.562 \\
$D_{adv} + D_{obj}$ & \textbf{20.596} & \textbf{11.609} & \textbf{15.108} & \textbf{6.797} & \textbf{18.478} & \textbf{10.113} & \textbf{17.986} & \textbf{9.421} \\
\bottomrule
\end{tabular}
\end{center}
\caption{Comparison of the use of $D_{adv}$ only, and the combination of $D_{adv}$ and $D_{obj}$ as in our design.}
\label{table:metrics_disc_ablation}
\end{table*}

\begin{figure*}[t]
\begin{center}

\resizebox{1.0\linewidth}{!}{

\begin{tabular}{c c c c c}
\includegraphics[width = 1.23in]{figures/additional_boxes/bed0753_bg.png} &
\includegraphics[width = 1.23in]{figures/additional_boxes/bed0753_bachgan.png} &
\includegraphics[width = 1.23in]{figures/additional_boxes/bed0753_spade.png} &
\includegraphics[width = 1.25in]{figures/additional_boxes/bed0753_layout.png} &
\includegraphics[width = 1.23in]{figures/additional_boxes/bed0753_nsd.jpg} \\
\includegraphics[width = 1.23in]{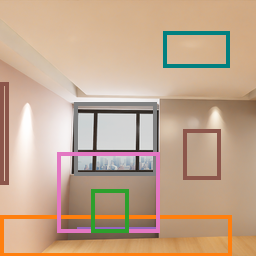} &
\includegraphics[width = 1.23in]{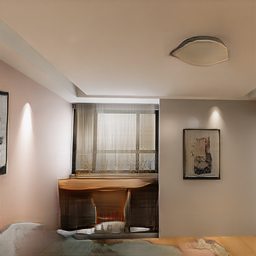} &
\includegraphics[width = 1.23in]{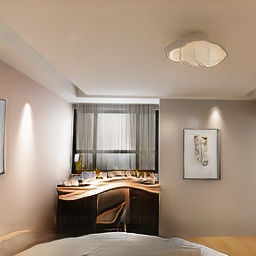} &
\includegraphics[width = 1.25in]{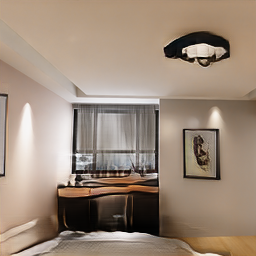} &
\includegraphics[width = 1.23in]{figures/baseline_boxes/bed3378_nsd.png} \\
\includegraphics[width = 1.23in]{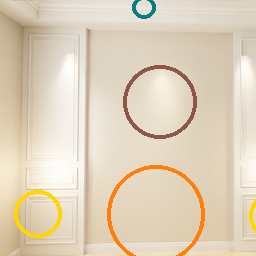} &
\includegraphics[width = 1.23in]{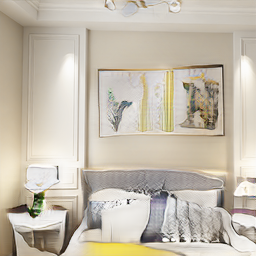} &
\includegraphics[width = 1.23in]{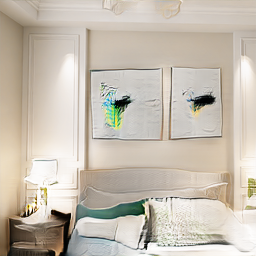} &
\includegraphics[width = 1.25in]{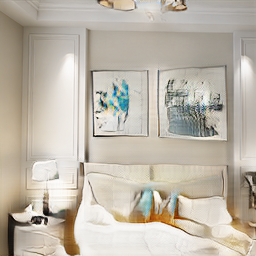} &
\includegraphics[width = 1.23in]{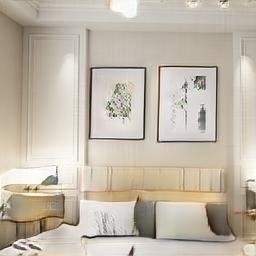} \\
\includegraphics[width = 1.23in]{figures/additional_points/liv1935_bg.png} &
\includegraphics[width = 1.23in]{figures/additional_points/liv1935_bachgan.png} &
\includegraphics[width = 1.23in]{figures/additional_points/liv1935_spade.png} &
\includegraphics[width = 1.25in]{figures/additional_points/liv1935_layout.png} &
\includegraphics[width = 1.23in]{figures/additional_points/liv1935_nsd.jpg} \\
Input & BachGAN & SPADE & He et al. & Ours \\

\multicolumn{5}{c}{
\includegraphics[width = 0.08in]{figures/palette/class_lamp.png} \small{lamp} \quad 
\includegraphics[width = 0.08in]{figures/palette/class_picture.png} \small{picture} \quad
\includegraphics[width = 0.08in]{figures/palette/class_bed.png} \small{bed} \quad 
\includegraphics[width = 0.08in]{figures/palette/class_curtain.png} \small{curtain} \quad
\includegraphics[width = 0.085in]{figures/palette/class_desk.png} \small{desk} \quad
\includegraphics[width = 0.08in]{figures/palette/class_chair.png} \small{chair} \quad 
\includegraphics[width = 0.08in]{figures/palette/class_nightstand.png} \small{nightstand} \quad
\includegraphics[width = 0.08in]{figures/palette/class_sofa.png} \small{sofa} \quad 
}\\

\end{tabular}

} 

\end{center}
\caption{Generation results of our method and other baselines, using box label format (the first two rows) and point label format (the last two rows). For point label format, the center and radius of each circle represent the location $\mathbf{c}_i$ and size $s_i$ of an object (see Eq.~(\ref{eq:point_label})). Best view with zoom.}
\label{fig:baseline_boxes}
\end{figure*}

\subsection{Quantitative Evaluation}

We quantitatively evaluate the image synthesis ability of our method using the Frechet Inception Distance (FID)~\cite{heusel2017gans} and the Kernel Inception Distance (KID)~\cite{binkowski2018demystifying}. Both FID and KID measure the dissimilarity between inception representations~\cite{Szegedy_CVPR15} of a synthesized output and its real version. Wasserstein distance and polynomial kernel are used in FID and KID respectively as dissimilarity metrics~\cite{obukhov2020torchfidelity}.

For quantitative evaluation purpose, we used pairs of background and decorated images from ground-truth. We also extracted bounding boxes and object masks of decorated objects from the ground-truth to construct box labels and point labels. We report the performance of our method and other baselines in Table~\ref{table:metrics_points}. As shown in the results, our method outperforms all the baselines on both bedroom and living room test sets, with both box label and point label format in FID metric. The same observation is true for the KID metric, with the only exception that He et al.'s method~\cite{he2021context} is ranked best with box label format for bedroom scenes.




Our method also has a computational advantage. Specifically, the BachGAN baseline took roughly the same amount of training time as our method but required four GPUs. In contrast, our method can produce even better results using less computational resources.

Table~\ref{table:metrics_points} also shows that point label scheme slightly outperforms box label scheme. However, as discussed in the next section, each scheme is favored to specific types of objects. For example, from a usage perspective, box label format has a strong focus on user's desire on how a decorated object appears, while point label format offers more flexibility and autonomy to the NSD system. 


\begin{figure*}[t]
\begin{center}
\resizebox{\linewidth}{!}{
\begin{tabular}{c c c c c c}
\includegraphics[width = 1.06in]{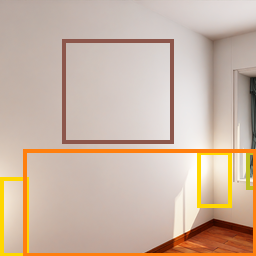} &
\includegraphics[width = 1.06in]{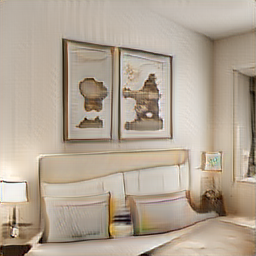} &
\includegraphics[width = 1.06in]{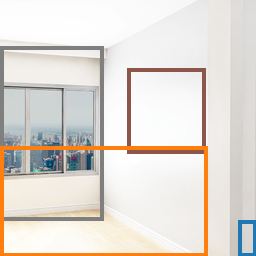} &
\includegraphics[width = 1.06in]{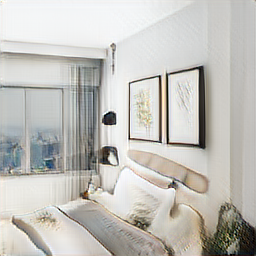} &
\includegraphics[width = 1.06in]{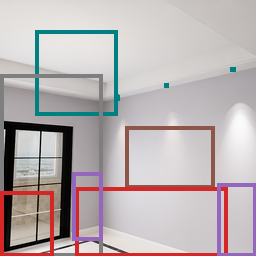} &
\includegraphics[width = 1.06in]{figures/boxes_vs_points/liv2825_boxes_fake.png} \\
\includegraphics[width = 1.06in]{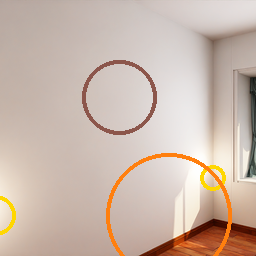} &
\includegraphics[width = 1.06in]{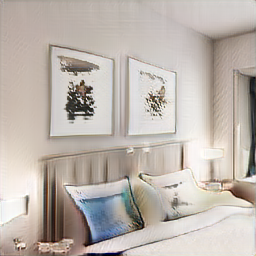} &
\includegraphics[width = 1.06in]{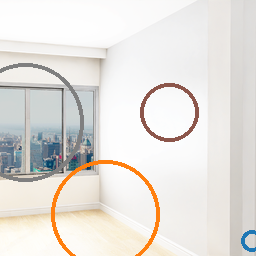} &
\includegraphics[width = 1.06in]{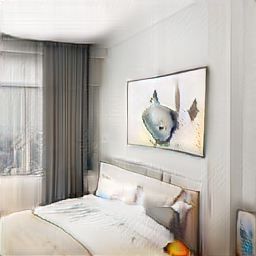} &
\includegraphics[width = 1.06in]{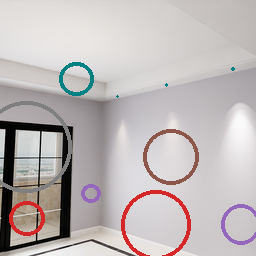} &
\includegraphics[width = 1.06in]{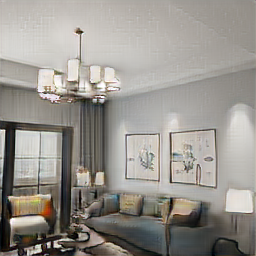} \\
Input & Ours & Input & Ours & Input & Ours \\

\multicolumn{6}{c}{
\includegraphics[width = 0.08in]{figures/palette/class_picture.png} \small{picture} \quad
\includegraphics[width = 0.08in]{figures/palette/class_bed.png} \small{bed} \quad 
\includegraphics[width = 0.08in]{figures/palette/class_nightstand.png} \small{nightstand} \quad
\includegraphics[width = 0.08in]{figures/palette/class_curtain.png} \small{curtain} \quad
\includegraphics[width = 0.08in]{figures/palette/class_cabinet.png} \small{cabinet} \quad
\includegraphics[width = 0.08in]{figures/palette/class_lamp.png} \small{lamp} \quad 
\includegraphics[width = 0.08in]{figures/palette/class_sofa.png} \small{sofa} \quad
\includegraphics[width = 0.08in]{figures/palette/class_table.png} \small{table} \quad
}\\
\end{tabular}
} 
\end{center}
\caption{Generation results (from the same input) using box label format (top row) and point label format (bottom row). While box label format suits small and relatively fixed-size objects, point label format is more flexible to describe large objects whose dimensions can be adjusted automatically.}
\label{fig:boxes_vs_points}
\end{figure*}

\subsection{Ablation Study}
In this experiment, we prove the role of the additional discriminator $D_{obj}$. Recall that $D_{obj}$ is branched off from $D_{adv}$ and combines $L$ at various scales (see Figure~\ref{fig:design}). To validate the role of $D_{obj}$, we amended the architecture of the discriminator $D$ by directly concatenating the object layout $L$ with the decorated image $Y$ to make the input for $D$, like the designs in~\cite{park2019SPADE} and \cite{li2020BachGAN}. Experimental results are in Table~\ref{table:metrics_disc_ablation}, which clearly confirms the superiority of our design for the discriminator (i.e., using both $D_{adv}$ and $D_{obj}$) over the use of $D_{adv}$ only.

\subsection{Qualitative Evaluation}

\paragraph{Generation quality.} We qualitatively compare our method with the baselines, on both box label and point label format in Figure~\ref{fig:baseline_boxes}. As shown in these results, our method is able to generate details in foreground objects. We hypothesize that our method successfully incorporates the layout information in early layers in the generator as each pixel in a down-scaled object layout encourages object generation within a specific local region in the output image. 

\paragraph{Box label vs. point label.} Figure~\ref{fig:boxes_vs_points} visualizes some generation results using box label and point label format on the same input background. In this experiment, on each scene, box labels and point labels were derived from the same set of objects. We observed that some object classes are better suited to a particular label format. For example, small objects and those whose aspect ratio can be varied (e.g., pictures can appear in either portrait or landscape shape) should be described using box label format. On the other hand, objects that often occupy large areas in a scene, such as beds and sofas, tend to have less distortion and clearer details when represented with point label format.

\begin{table}[t]
\begin{center}
\begin{tabular}{l | c c | c c } 
\toprule
& \multicolumn{2}{c}{Bedroom} & \multicolumn{2}{c}{Living room} \\
\midrule
Method & FID & KID{\tiny$\times10^3$} & FID & KID{\tiny$\times10^3$} \\
\midrule
SPADE~\cite{park2019SPADE} & 22.985 & 10.993 & 23.630 & 13.620 \\
BachGAN~\cite{li2020BachGAN} & 19.914 & 8.119 & 23.391 & 13.006 \\
Ours & \textbf{17.489} & \textbf{8.007} & \textbf{21.019} & \textbf{11.277} \\
\bottomrule
\end{tabular}
\end{center}

\caption{Performance of our method and the baselines using default object sizes. Lower FID/KID values indicate better performance.}
\label{table:metrics_default}
\end{table}

\begin{figure*}[t]
\begin{center}
\resizebox{0.9\linewidth}{!}{
\begin{tabular}{c c c c}
\includegraphics[width = 1.23in]{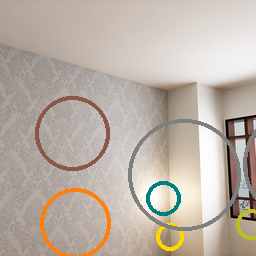} &
\includegraphics[width = 1.23in]{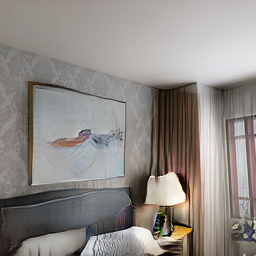} &
\includegraphics[width = 1.23in]{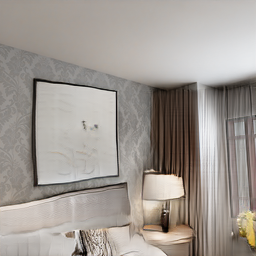} &
\includegraphics[width = 1.23in]{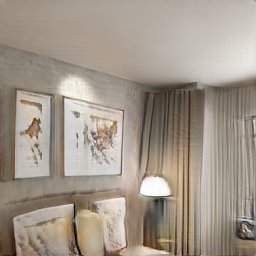} \\
\includegraphics[width = 1.23in]{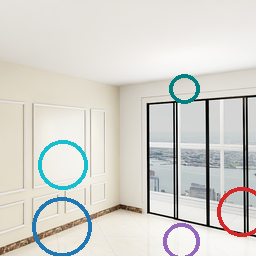} &
\includegraphics[width = 1.23in]{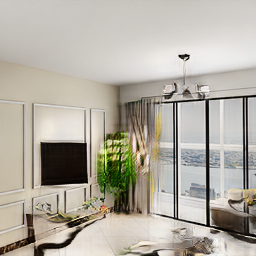} &
\includegraphics[width = 1.23in]{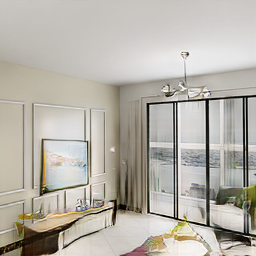} &
\includegraphics[width = 1.23in]{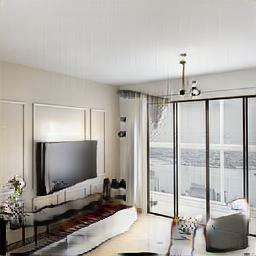} \\
Input & BachGAN & SPADE & Ours \\

\multicolumn{4}{c}{
\includegraphics[width = 0.08in]{figures/palette/class_cabinet.png} \small{cabinet} \quad 
\includegraphics[width = 0.08in]{figures/palette/class_lamp.png} \small{lamp} \quad 
\includegraphics[width = 0.08in]{figures/palette/class_picture.png} \small{picture} \quad
\includegraphics[width = 0.08in]{figures/palette/class_bed.png} \small{bed} \quad 
\includegraphics[width = 0.08in]{figures/palette/class_curtain.png} \small{curtain}
}\\
\multicolumn{4}{c}{
\includegraphics[width = 0.08in]{figures/palette/class_sofa.png} \small{sofa} \quad
\includegraphics[width = 0.08in]{figures/palette/class_table.png} \small{table} \quad
\includegraphics[width = 0.08in]{figures/palette/class_television.png} \small{television} \quad 
\includegraphics[width = 0.08in]{figures/palette/class_nightstand.png} \small{nightstand} \quad
\includegraphics[width = 0.08in]{figures/palette/class_8.png} \small{pillow}
}
\end{tabular}
} 
\end{center}
\caption{Generation results using default values for object size.}
\label{fig:baseline_default}
\end{figure*}

\subsection{Setting Object Sizes}
\label{section:object_size}
In the quantitative assessment, the sizes of decorated objects in the point label format (i.e., $s_i$ in Eq.~(\ref{eq:point_label})) were retrieved from ground-truth. However, in reality, this information is provided by the user. In this experiment, we investigate a simpler input for the point label format where the object sizes are set by default values rather than given by either the ground-truth or user. In particular, for each decorated object $o_i$, we set the size $s_i$ to the median size of all the objects having the same object class with $o_i$ in the training data. Particularly, the ground truth value of $s_i$ of each object is given by $s_i=m\sqrt{A_i}$, where $m$ is a fixed constant and $A_i$ is the area (i.e., the no. of pixels) of the object mask of $o_i$. Here we set $m=2.5$ for our experiments. 

\begin{table}[t]
\footnotesize
\begin{center}
\begin{tabular}{l | c c | c c } 
\toprule
& \multicolumn{2}{c}{Bedroom} & \multicolumn{2}{c}{Living room} \\
\midrule
Method & FID & KID{\tiny$\times10^3$} & FID & KID{\tiny$\times10^3$} \\
\midrule
Class mean & 20.145 & 10.457 & 27.365 & 14.639 \\
Class median & 17.489 & 8.007 & 21.019 & 11.277 \\
\midrule
$m=4.0$ & 24.045 & 14.701 & 28.128 & 15.979 \\
$m=2.5$(*) & \textbf{15.108} & \textbf{6.797} & \textbf{17.986} & \textbf{9.421} \\
$m=1.0$ & 20.201 & 9.105 & 19.114 & 11.797 \\
\bottomrule
\end{tabular}
\end{center}
\caption{Performance using different object sizes methods. (*) indicate the default setting used in our experiments.}
\label{table:points_ablation}
\end{table}

We applied this setting to all the baselines and reported their performance in Table~\ref{table:metrics_default}. We observed that compared with using ground-truth values for the object sizes (see Table~\ref{table:metrics_points}), setting the object sizes to default values degrades the performance of all the competitors. However, our method still consistently outperforms all the baselines on all the test sets, using both FID and KID metrics. We illustrate several results of this setting in Figure~\ref{fig:baseline_default}.
In Table~\ref{table:points_ablation}, we further compare different variants of point label format, including the use of mean value and median value to compute $s_i$. We show the results of using alternate values for $m$ in Table~\ref{table:points_ablation}, which clearly confirms our choice for $m$ for the best performance.

\begin{figure}[t]
\begin{center}
\begin{tabular}{c c c c c c}
\includegraphics[width = 0.75in]{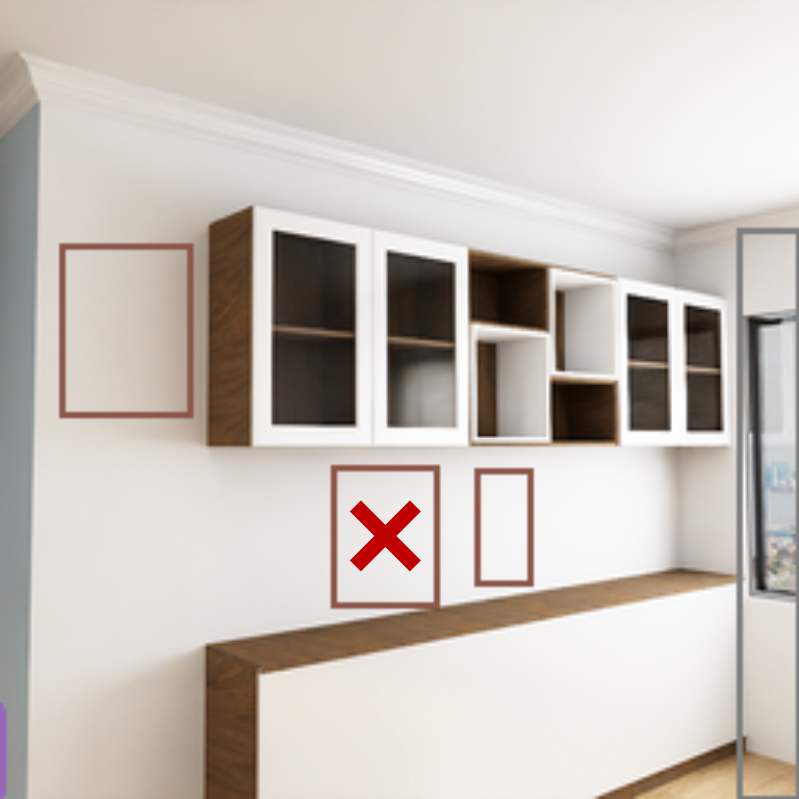} &
\includegraphics[width = 0.75in]{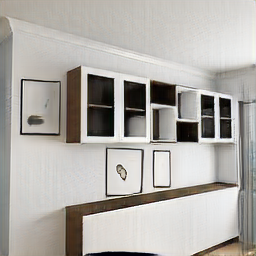} &
\includegraphics[width = 0.75in]{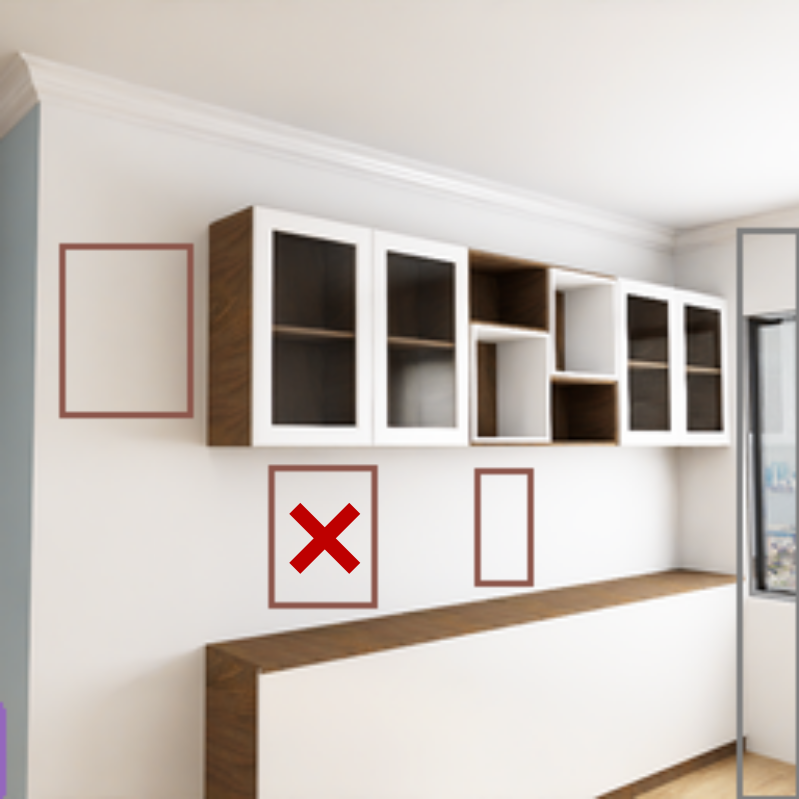} &
\includegraphics[width = 0.75in]{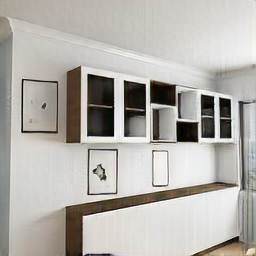} &
\includegraphics[width = 0.75in]{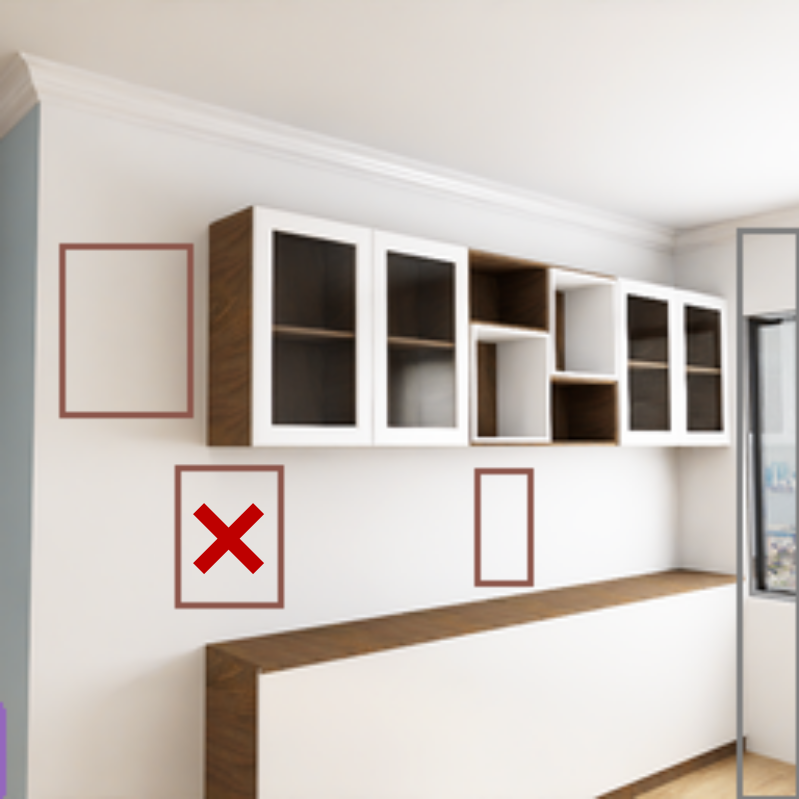} &
\includegraphics[width = 0.75in]{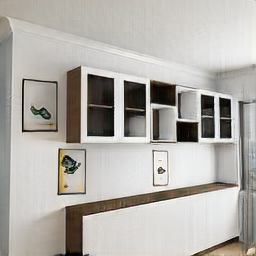} \\
\includegraphics[width = 0.75in]{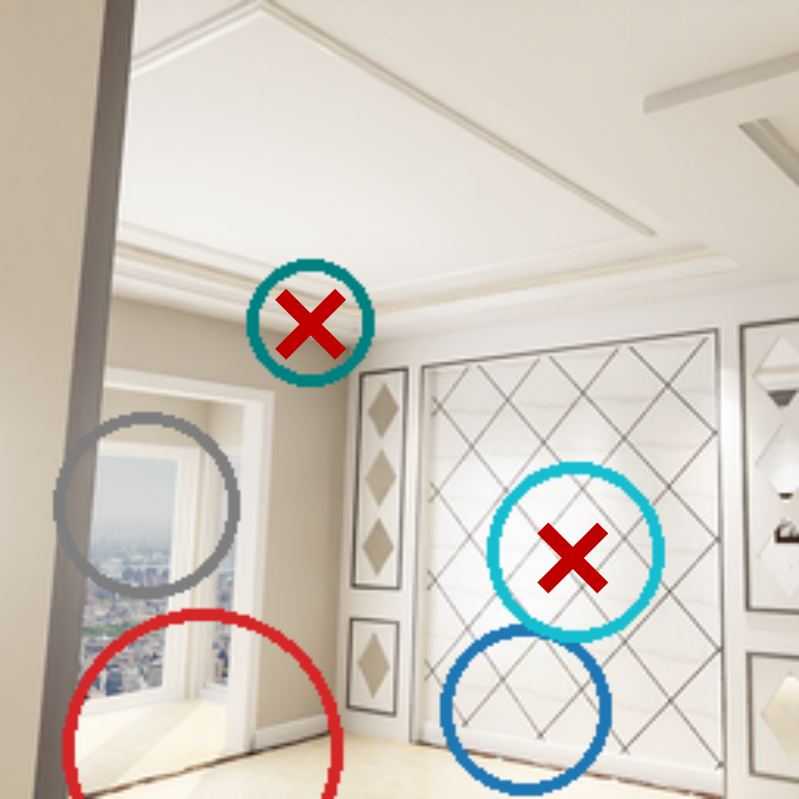} &
\includegraphics[width = 0.75in]{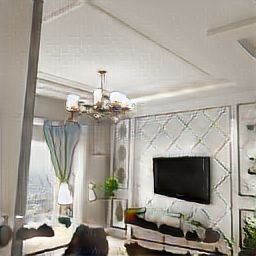} &
\includegraphics[width = 0.75in]{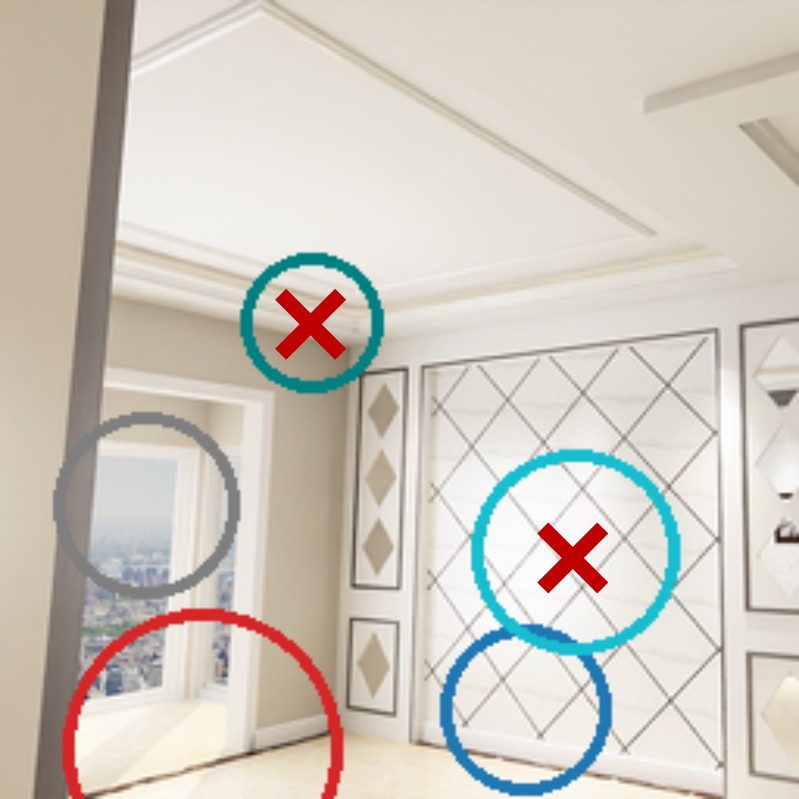} &
\includegraphics[width = 0.75in]{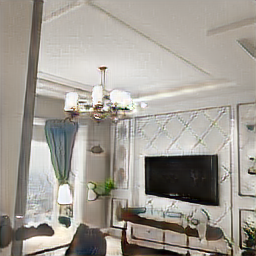} &
\includegraphics[width = 0.75in]{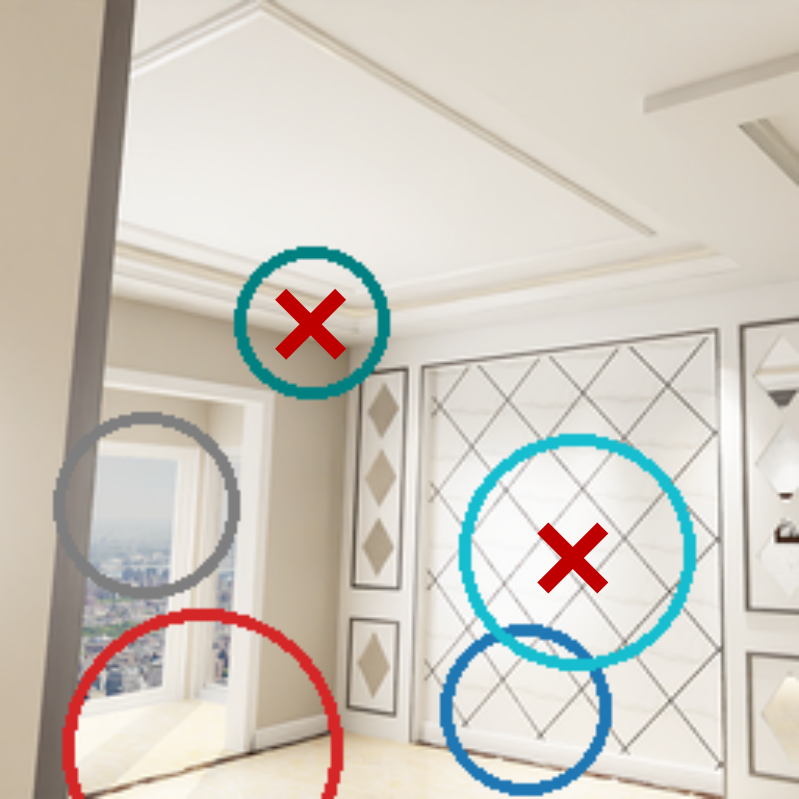} &
\includegraphics[width = 0.75in]{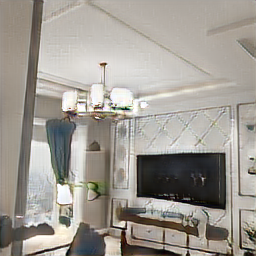} \\
\multicolumn{6}{c}{
\includegraphics[width = 0.08in]{figures/palette/class_picture.png} \small{picture} \space\space 
\includegraphics[width = 0.08in]{figures/palette/class_lamp.png} \small{lamp} \space\space 
\includegraphics[width = 0.08in]{figures/palette/class_curtain.png} \small{curtain} \space\space
\includegraphics[width = 0.08in]{figures/palette/class_sofa.png} \small{sofa} \space\space
\includegraphics[width = 0.08in]{figures/palette/class_television.png} \small{television} \space\space 
\includegraphics[width = 0.08in]{figures/palette/class_cabinet.png} \small{cabinet}
}
\end{tabular}
\end{center}
\caption{Generation results under different settings for objects' locations and sizes. Manipulated objects are marked with ``X''. Top two rows: we change the location of a painting by moving its bounding box towards the left. Bottom two rows: we adjust the size of a ceiling light and a TV by changing the radius at their centers.} 
\label{fig:move_labels}
\end{figure}

\begin{figure}[t]
\begin{center}
\begin{tabular}{c c c c}
\includegraphics[width = 1.15in]{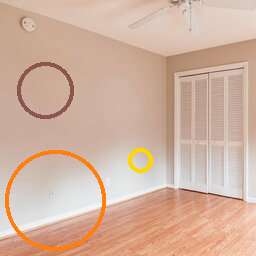} &
\includegraphics[width = 1.15in]{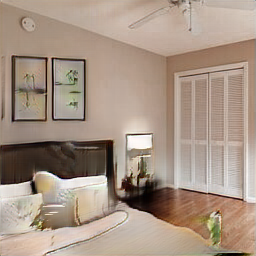} &
\includegraphics[width = 1.15in]{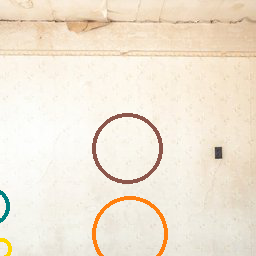} &
\includegraphics[width = 1.15in]{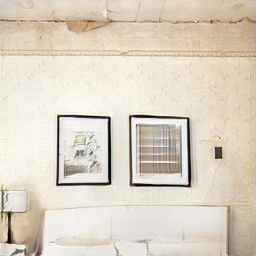} \\
Background & Generated & Background & Generated \\
\multicolumn{4}{c}{
\includegraphics[width = 0.08in]{figures/palette/class_picture.png} \small{picture} \quad
\includegraphics[width = 0.08in]{figures/palette/class_bed.png} \small{bed} \quad 
\includegraphics[width = 0.08in]{figures/palette/class_nightstand.png} \small{nightstand} \quad
\includegraphics[width = 0.08in]{figures/palette/class_lamp.png} \small{lamp} \quad 
}
\end{tabular}
\end{center}
\caption{Generation results on real-world scenes.}
\label{fig:real_rooms}
\end{figure}

\subsection{Layout Manipulation}
We demonstrate in Figure~\ref{fig:move_labels} how our method performs under various settings for decorated objects' positions and sizes. For box label format, these settings can be done by changing the location and dimensions of objects' bounding boxes. For point label format, we adjust the center position and radius of objects. As shown in Figure~\ref{fig:move_labels}, our method can generate contents adaptively to different settings, proving its applicability in creating conceptual designs of interior space.

Please also refer to our supplementary material for additional discussions on the diversity of the generation results, the impact of the background to furniture, and iterative decoration, where the furniture can be added one by one.

\subsection{Generalization to Real-World Data} 
To validate the universality of our method, we experimented with it on several real-world scenes not included in the Structured3D dataset. We present several results of this experiment in Figure~\ref{fig:real_rooms}. Interestingly, we discovered that paintings and furniture are generated in different artistic styles but still in line with the entire decoration and background style.

\subsection{User study}
\label{sec:user_study}

\begin{figure}[t]
\begin{center}

\begin{minipage}{0.35\linewidth}
\includegraphics[width=\linewidth]{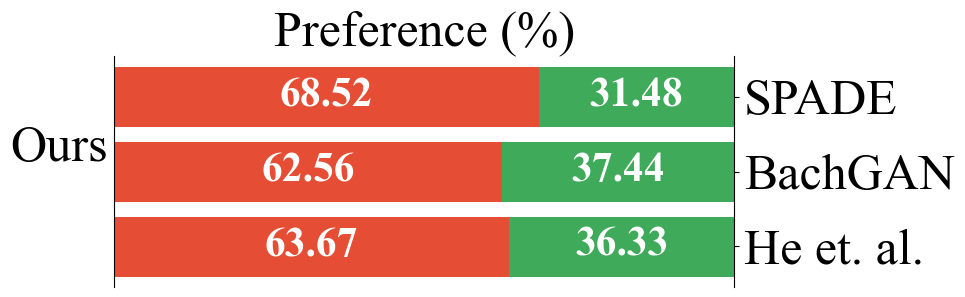}\\ 
\centerline{(a)}
\end{minipage}
\begin{minipage}{0.30\linewidth}
\includegraphics[width=\linewidth]{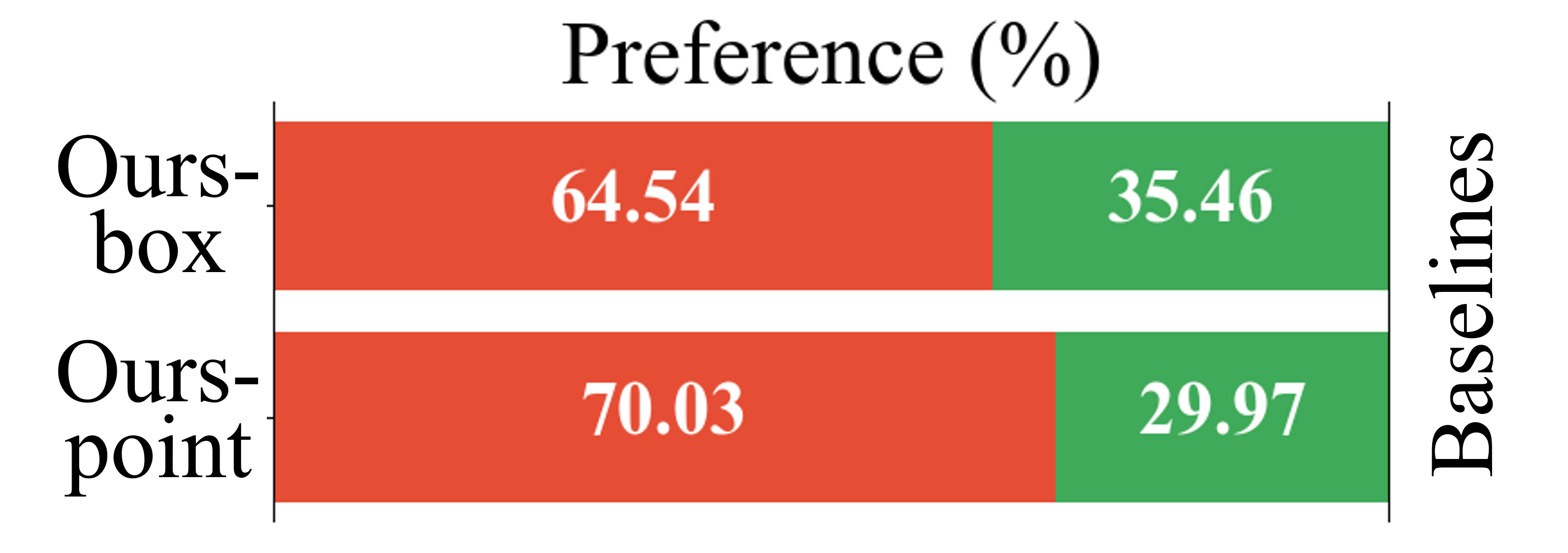}
\centerline{(b)}
\end{minipage}
\begin{minipage}{0.30\linewidth}
\includegraphics[width=\linewidth]{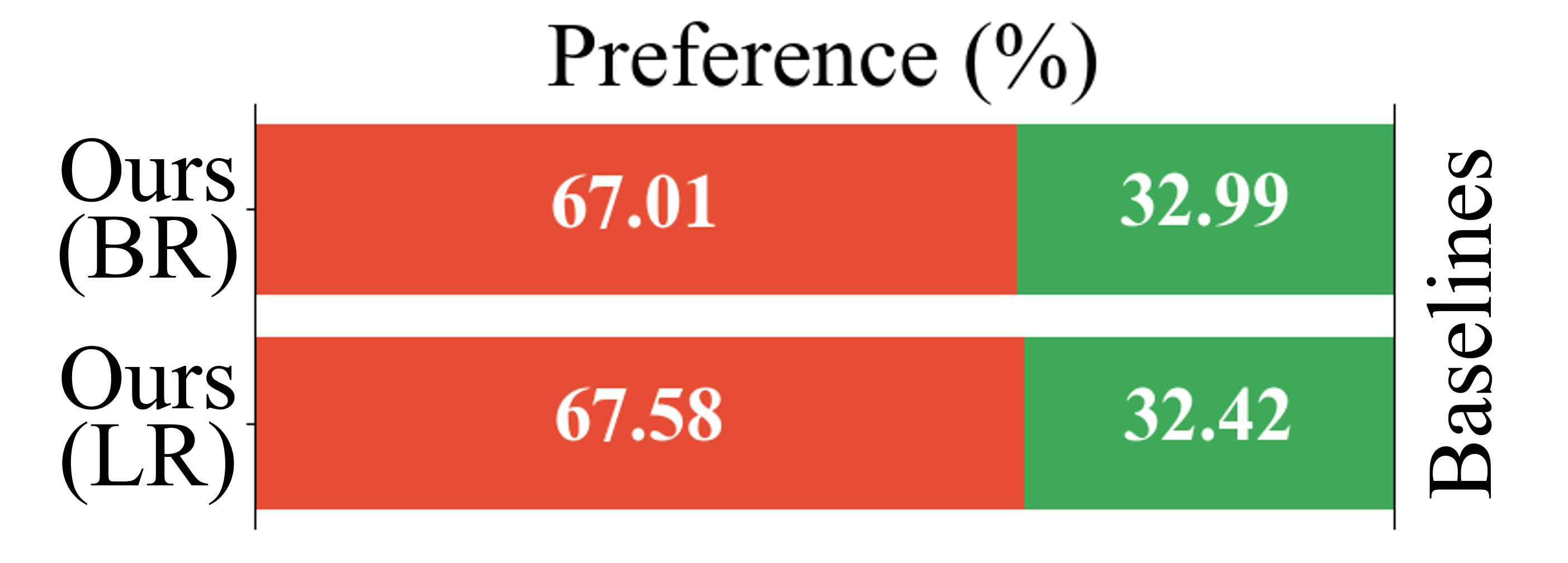}\\ 
\centerline{(c)} 
\end{minipage}

\end{center}
\caption{User study's results: Preference with (a) different methods, (b) box labels vs point labels, and (c) different room types. (BR = bedroom, LR = living room)}
\label{fig:study_results}
\end{figure}

In addition to quantitative evaluation, we conducted a user study on the generation quality of our method and other baselines. 
We collected responses from 26 participants and summarized these results in Figure~\ref{fig:study_results}. I general, our model is often preferred in generating images with point label format, especially in the bedroom test case with fewer objects and clutter. When using box label format, our method still produces results around the same level of quality compared with the baselines. More details of our user study is in the supplementary material.

%% file: sections/conclusion.tex
\section{Conclusion}

We introduced a new task called neural scene decoration. The task aims to render an empty indoor space with furniture and decorations specified in a layout map. To realize this task, we propose an architecture conditioned on a background image and an object layout map where decorated objects are described via either bounding boxes or rough locations and sizes. We demonstrate the capability of our method in scene design over previous works on the Structured3D dataset. 
Neural scene decoration is henceforth a step toward building the next generation of user-friendly interior design and rendering applications. 
Future work may include better support of sequential object generation~\cite{turkoglu2019layer}, interactive scene decoration, and integration of more advanced network architecture.
